\documentclass[10pt,twocolumn,letterpaper]{article}

\usepackage[pagenumbers]{cvpr}

\usepackage{booktabs}

\definecolor{cvprblue}{rgb}{0.21,0.49,0.74}
\usepackage[pagebackref,breaklinks,colorlinks,allcolors=cvprblue]{hyperref}

\title{Match the Positive to the Index: Why Whole-Tile Training Limits Dense Cross-View Geo-Localization}

\author{
Fabian Deuser$^{1,3}$ \quad Tobias Hallmen$^{2}$ \quad Jasmin Steinbrecher$^{3}$ \quad Martin Werner$^{3,4}$ \quad Norbert Oswald$^{1}$ \\[0.5em]
\parbox{0.95\textwidth}{\centering\small
$^{1}$Institute of Distributed Intelligent Systems (ETTI~2), University of the Bundeswehr Munich, Neubiberg, Germany \\
$^{2}$University of Augsburg, Augsburg, Germany \\
$^{3}$Technical University of Munich, Germany; TUM School of Engineering and Design, Department of Aerospace and Geodesy \\
$^{4}$TUM School of Computation, Information and Technology; Munich Data Science Institute (MDSI)
}
}

\begin{document}
\maketitle
\begin{abstract}
Cross-view geo-localization retrieves georeferenced overhead tiles matching a ground-level query. Densely indexing overlapping, tile-sized crops provides more precisely aligned references but introduces visually similar candidates. In line with observations from previous work, we show that this can reduce R@1 by up to 12.8 percentage points for published models trained with whole-tile positives. We attribute this to a mismatch between the training positive and the dense index. Whole-tile training treats an overhead tile as a positive match regardless of the camera's position within it, providing no incentive to rank geographically closer references higher. Image content then decides among the crops near the camera, and the top-ranked crop often belongs to a neighboring tile, which counts as a miss. A training-free spatial vote over the crops more than recovers the lost R@1 and raises the same models up to 10.8 percentage points above their R@1 without the dense index. To address the mismatch itself, we introduce LaSo (\emph{La}ttice positives with \emph{So}ft siblings), which uses the nearest lattice crop as the positive and neighboring crops as distance-weighted soft targets. Using one embedding per crop and no re-ranking, LaSo outperforms spatially voted whole-tile models by 7.5--33.3 percentage points in R@1 across VIGOR and DReSS, achieving median localization errors as low as 1.3\,m on VIGOR and 2.6\,m on DReSS. 
\end{abstract}
    
\section{Introduction}
\label{sec:intro}
\begin{figure}[t]
    \centering
    \includegraphics[width=0.9\linewidth]{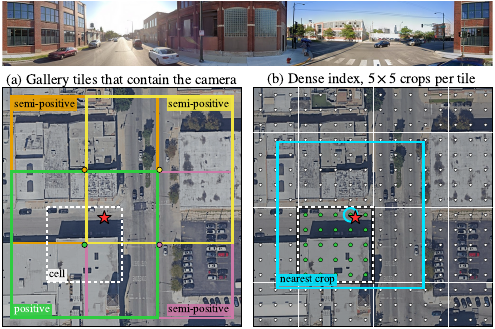}
    \caption{\textbf{Standard gallery and dense index} for one VIGOR query (top, camera at the star). (a)~Overlapping gallery tiles: the positive (green), whose cell (dashed) contains the camera, and three semi-positives. (b)~Dense index: 25 crops per tile, the tile itself and 24 shifted crops, with centers (dots) spread over its cell (white boundaries, green: crops of the positive). Cyan: the crop closest to the camera.}
    \label{fig:setting}
\end{figure}
Cross-view geo-localization matches ground-level images against a gallery of georeferenced overhead tiles~\cite{lin2013cross}. While CVUSA~\cite{workman2015wide,zhai2017predicting}, CVACT~\cite{liu2019lending}, and CV-Cities~\cite{huang2024cv} pair each query with a camera-centered tile, VIGOR~\cite{zhu2021vigor} and DReSS~\cite{xia2025cross} use query-independent grids of overlapping tiles (\cref{fig:setting}a). The \emph{positive} of a query is the tile whose center is nearest to the camera. Other tiles that also contain the camera have more distant centers and count as misses (semi-positives). All camera positions that share the same positive form its \emph{cell}, a square around the tile center. Seen from the camera, the center of the positive can therefore lie anywhere within a square of the same size around the camera. Retrieving the correct tile thus provides only coarse localization, typically within tens of meters.

One way to localize more precisely is to increase the gallery's spatial resolution. Around each tile, we add overlapping, shifted crops of the same size whose centers are distributed across its cell (\cref{fig:setting}b). We call the result a \emph{dense index} and the original tiles alone the \emph{standard gallery}. The crop centers form a lattice finer than the tile grid, placing multiple references within the square around the camera. Since the closest crop always belongs to the positive tile (cyan in \cref{fig:setting}b), a model favoring geographically closer references should benefit from the dense index.

However, previous studies suggest otherwise. \citet{shi2022beyond} observed no consistent gain from denser reference sampling and suspected that nearby references are too similar to be told apart, and \citet{shore2025peng} argue that overlapping references make incorrect retrievals more likely. On a dense index, Sample4Geo~\cite{deuser2023sample4geo} on both benchmarks, the retrieval stage of UnifyGeo~\cite{song2025unified} on VIGOR, and AuxGeo~\cite{xia2025cross} on DReSS all lose up to 12.8 percentage points of R@1, the share of queries whose top-ranked reference belongs to the positive tile. Yet median localization error decreases, and far-off retrievals become less frequent, even as the top-ranked crop more often belongs to a neighboring tile and thus counts as a miss (\cref{sec:drop}). The models thus find the right neighborhood but do not rank the crops within it by their distance to the camera.

We trace this behavior to a mismatch between the training positive and the dense index. Nearby references do look alike, as \citet{shi2022beyond} suspected, but a model can tell them apart when training requires it. Current models, which we call \emph{whole-tile models}, are instead trained with the whole tile as the positive. The center of this positive lies anywhere within the square around the camera, and the centers of all other tiles lie outside it. The objective thus rewards high similarity anywhere in this square and gives no incentive to rank closer references higher. Consequently, the learned similarity is, on average, nearly constant over much of the square and forms a \emph{plateau} rather than a peak at the camera (\cref{fig:score_field}). When we confine the training positive to a smaller square, the plateau shrinks with it (\cref{sec:plateau}).

The standard gallery hides the plateau because only one tile center lies within the square, that of the positive. The dense index places multiple crops there, some of them from neighboring tiles, and image content rather than distance to the camera decides their ranking. A training-free spatial vote exploits that the most similar crops cluster around the camera, and raises every tested whole-tile model above its standard-gallery R@1, but it only reads out the plateau and cannot change the underlying similarities.

To shrink the plateau, we introduce \textbf{LaSo} (\textbf{La}ttice positives with \textbf{So}ft siblings), which selects the lattice crop nearest to the camera as the training positive. This restricts the positive to a square whose side is the lattice stride, the spacing between neighboring crops, thereby aligning it with the dense index and narrowing the plateau to a peak around the camera (\cref{fig:score_field}). LaSo further incorporates neighboring crops as soft targets weighted by their distance to the camera. Our contributions are:
\begin{itemize}
    \item An analysis that links the R@1 degradation on a dense index to whole-tile training, with measurements in 14 model--split cases (\cref{sec:analysis}).
    \item A training-free spatial vote that recovers the R@1 loss and improves retrieval beyond the standard-gallery baseline (\cref{sec:vote}).
    \item LaSo, which improves R@1 by 7.5--33.3 percentage points over spatially voted whole-tile models across VIGOR and DReSS and localizes the camera to within a few meters without a separate localization model (\cref{sec:method,sec:experiments}).
\end{itemize}

\section{Related Work}
\label{sec:related}
\paragraph{Cross-view geo-localization.}
Early approaches matched hand-crafted features across ground-level and overhead imagery~\cite{lin2013cross} before Siamese networks learned a shared embedding space~\cite{lin2015learning,workman2015wide}. Subsequent work improved feature aggregation~\cite{hu2018cvmnet,shi2019safa,shi2020optimal}, adopted transformers~\cite{yang2021cross,zhu2022transgeo,zhang2023cross}, or reduced the viewpoint gap through polar transformations, view synthesis, or bird’s-eye-view representations~\cite{shi2019safa,regmi2018cross,regmi2019bridging,toker2021coming,ye2024cross}. Sample4Geo~\cite{deuser2023sample4geo} showed that a pre-trained ConvNeXt combined with a symmetric InfoNCE loss~\cite{oord2018representation} and hard-negative mining can outperform many task-specific architectures. Recent approaches build on it with large-scale pre-training~\cite{huang2024cv,geobridge2026,geobridgepp2025,zhang2026geo} or additional modules~\cite{xia2025cross,song2026warp}, or add a second stage that re-ranks the retrieved tiles~\cite{fervers2023c,deuser2024kmeansrerank,song2025unified}. On the grid benchmarks, all of them train with the whole tile as the positive and retrieve on the standard gallery. VIGOR~\cite{zhu2021vigor} additionally ties the similarity of each semi-positive to that of the positive by the ratio of their overlaps with a camera-centered tile, and regresses the offset of the camera within the retrieved tile. Both act on the tiles of the standard gallery and leave the whole tile as the positive, whereas LaSo weights the crops of the dense index by their distance to the camera.

\paragraph{Dense references and soft targets.}
Retrieval over densely sampled references predates the grid-based benchmarks. \citet{hu2020image} match ground-level images against a lattice of overhead references spaced 5\,m apart, with a model pre-trained on camera-centered CVUSA pairs. Several works move the positive or use soft targets. \citet{xia2021cross} offset the positive from the camera, WAG~\cite{downes2022city} adds a semi-positive term for queries far from the tile center to permit a coarser tile grid, and Game4Loc~\cite{ji2025game4loc} weights the InfoNCE target of partially overlapping drone--overhead pairs by their overlap. Statewide geolocalization~\cite{fervers2024statewide} also offsets the positive but masks nearby references from the negatives, whereas LaSo takes the positive from the lattice of the index and trains its neighbors as soft targets. None of these works explain why whole-tile training degrades retrieval on a dense index. We trace this to the learned similarity and remove it by matching the training positive to the index.
\paragraph{Fine-grained localization.}
VIGOR is also used for fine-grained localization, which estimates the camera position within a given tile~\cite{lentsch2023slicematch,xia2023convolutional,wang2023fine,xia2025fg}. LaSo is not given the tile but retrieves over the whole dense index and takes the position from the retrieved crop (\cref{sec:main}). CVUSA and CVACT pair each query with a camera-centered tile and are already close to saturation, so we evaluate on VIGOR and DReSS, whose tile grids are independent of the queries.

\section{Grid Benchmarks and the Dense Index}
\label{sec:setting}
\subsection{Grid and notation}
VIGOR and DReSS arrange tile centers on regular grids. We denote the spacing between neighboring tile centers, the \emph{grid step}, by $g$. On VIGOR, $g=36.5$\,m (approximately half a tile width), and on DReSS, $g$ is $87.5\%$ of a tile width ($73$--$108$\,m, depending on city latitude; \cref{sec:supp_cell}). A query $q$ has a GPS camera position $\mathbf{x}\in\mathbb{R}^2$ in meters, unrestricted within the covered area, whereas tile centers $\mathbf{c}_t$ lie on the grid. Models embed queries and references and rank references by embedding cosine similarity $s(q,r)$, hereafter referred to as similarity. For any reference $r$, whether a tile or a crop introduced below, we define its camera-relative offset as $\boldsymbol{\delta}_r=\mathbf{c}_r-\mathbf{x}$, where $\mathbf{c}_r$ is its center. Both $\mathbf{x}$ and $\boldsymbol{\delta}_r$ depend on the query, and we omit the index $q$, as every quantity below refers to one query.

The positive is the tile whose center is nearest to the camera. On a square grid, this is the tile with $\|\boldsymbol{\delta}_t\|_\infty\leq g/2$, \ie, the tile whose center lies within the square $\mathcal{S}$ of side $g$ centered at $\mathbf{x}$. The cell of tile $t$ is the square of side $g$ centered at $\mathbf{c}_t$.
Test cameras are nearly uniformly distributed within their cells (\cref{fig:supp_cell}), making the positive-tile offset approximately uniform over $\mathcal{S}$.
We retain the official labels, which agree with this definition for at least $99\%$ of test queries.
Both benchmarks provide same-area and cross-area splits, the latter holding out test cities.

\subsection{Dense index}
For each tile $t$, we define shifted references $r$ arranged in an $n\times n$ lattice of crops with centers
\begin{equation}
    \mathbf{c}_r \mathrel{:}= \mathbf{c}_t + \frac{g}{n}\boldsymbol{\nu}, \quad \boldsymbol{\nu} \in \left\{ -\frac{n-1}{2}, \ldots, \frac{n-1}{2} \right\}^2, \label{eq:lattice}
\end{equation}
which lie within the cell of $t$ (\cref{fig:setting}b). These references cover the same area as a tile, so the model sees them at the scale it was trained on. Smaller references would change that: \citet{fan2026one} report that Sample4Geo falls from 52.1 to 10.0 R@1 on their VIGOR-M benchmark when the gallery tiles cover half the ground width of the training tiles. The crops are extracted from a mosaic of gallery tiles stitched together at their georeferenced positions. A crop \emph{belongs to} the tile in whose cell its center lies, although it may also cover parts of neighboring tiles. We use grid exclusively for the tile centers and lattice for the crop centers. The crops from all tiles form a single lattice with stride $g/n$. For odd $n$, the central crop coincides with the original tile, so the dense index retains the standard gallery and requires no additional imagery. We use $n=5$, which yields a stride of 7.3\,m on VIGOR and 15--22\,m on DReSS and up to 10.6\,M crops (DReSS same-area); \cref{sec:ablation} compares other lattices.

\subsection{Protocol}
\label{sec:protocol}
We embed crops in the same way as tiles and retrieve the most similar crop through a single search over all crops. The \emph{best crop} is the top-ranked one, \ie, the crop with the highest similarity, which need not be the one nearest to the camera. The tile it belongs to is the retrieved tile. We rank tiles by the maximum similarity among their crops and define R@$k$ as the proportion of queries for which the positive tile ranks among the top $k$. Thus, the standard gallery and dense index share the same queries, imagery, and evaluation criterion: a retrieval is correct if the camera lies within the cell of the retrieved tile. The center of the retrieved reference also provides an estimate of the camera position. We define the localization error as its distance from the camera and compute it over all queries, including misses. R@1 thus tells whether the best crop belongs to the positive, and the localization error how far it lies from the camera. A single miss can lie kilometers away and would dominate a mean, so we report the median localization error throughout.
\begin{table}[t]
    \centering
    \small
    \setlength{\tabcolsep}{3pt}
    \resizebox{\columnwidth}{!}{%
    \begin{tabular}{@{}llcccccc@{}}
        \toprule
        & & \multicolumn{3}{c}{R@1 $\uparrow$} & \multicolumn{3}{c}{Err.\,[m] $\downarrow$} \\
        \cmidrule(lr){3-5}\cmidrule(lr){6-8}
        Split & Model & Tiles & Dense & +Vote & Tiles & Dense & +Vote \\
        \midrule
        VIGOR cross & Sample4Geo~\cite{deuser2023sample4geo} & 61.38 & 53.18 & 66.06 & 17.8 & 13.0 & 6.5 \\
                    & UnifyGeo~\cite{song2025unified}        & 61.25 & 51.69 & 66.12 & 17.9 & 13.8 & \textbf{6.3} \\
                    & Whole tile (ours)                      & \textbf{64.32} & \textbf{57.42} & \textbf{70.75} & \textbf{17.3} & \textbf{12.0} & \textbf{6.3} \\
        \midrule
        VIGOR same  & Sample4Geo~\cite{deuser2023sample4geo} & 77.88 & 66.89 & 80.55 & \textbf{15.7} & 10.0 & 5.0 \\
                    & UnifyGeo~\cite{song2025unified}        & \textbf{78.47} & 65.70 & \textbf{80.94} & \textbf{15.7} & 10.7 & \textbf{4.8} \\
                    & Whole tile (ours)                      & 76.96 & \textbf{67.88} & 80.65 & 15.8 & \textbf{9.5} & 5.1 \\
        \midrule
        DReSS cross & Sample4Geo~\cite{deuser2023sample4geo} & 29.29 & 28.65 & 36.39 & 2\,600 & 470 & 242 \\
                    & AuxGeo~\cite{xia2025cross}             & 32.27 & 29.11 & 39.38 & 1\,830 & 245 & 104 \\
                    & Whole tile (ours)                      & \textbf{33.37} & \textbf{32.96} & \textbf{42.99} & \textbf{1\,364} & \textbf{68.7} & \textbf{39.4} \\
        \midrule
        DReSS same  & Sample4Geo~\cite{deuser2023sample4geo} & 52.05 & \textbf{50.54} & 61.29 & 51.8 & \textbf{32.8} & 18.6 \\
                    & AuxGeo~\cite{xia2025cross}             & \textbf{54.37} & 47.83 & \textbf{62.77} & \textbf{49.7} & 35.6 & \textbf{17.2} \\
                    & Whole tile (ours)                      & 51.03 & 49.48 & 61.83 & 52.8 & 33.4 & 18.5 \\
        \bottomrule
    \end{tabular}}
    \caption{\textbf{R@1 and median localization error (Err.) of whole-tile models} on the standard gallery (Tiles), on the $5\times5$ dense index with the best crop (Dense), and with the spatial vote of \cref{sec:vote} (+Vote). UnifyGeo: its retrieval stage, without the re-ranking of \cref{tab:main}. Best values per split in bold.}
    \label{tab:dense_drop}
\end{table}

\begin{figure*}[t]
    \centering
    \includegraphics[width=\linewidth]{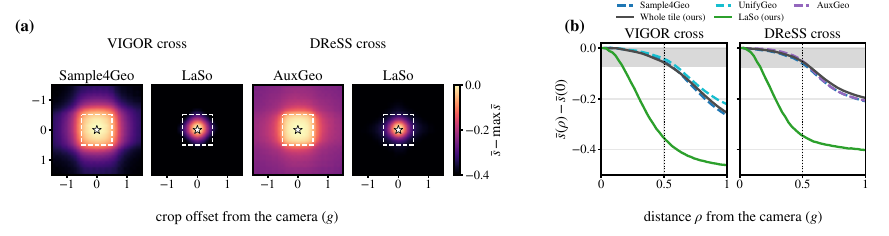}
    \caption{\textbf{The plateau is observed in every whole-tile model, and LaSo narrows it.} (a)~Mean similarity $\bar s$ of the crops by their offset from the camera, which lies at the star for every query, relative to its maximum, in bins of $0.1g$ over 5\,000 test queries, for Sample4Geo and LaSo on VIGOR cross and for AuxGeo and LaSo on DReSS cross. Dashed: the square $\mathcal{S}$, on which \cref{eq:plateau} predicts a plateau for the whole-tile positive. (b)~Mean similarity by the distance $\rho$ from the camera, $\bar s(\rho) - \bar s(0)$, in bins of $0.05g$ ($0.025g$ for LaSo), for the published models, our whole-tile model and LaSo. Gray band: the mean standard deviation $\sigma$ of the residual of the whole-tile models. Dotted: the nearest border of $\mathcal{S}$, at $0.5g$. Our whole-tile model and LaSo: ViT-B with public initialization.}
    \label{fig:score_field}
\end{figure*}

\section{Whole-Tile Models Lose on a Dense Index}
\label{sec:analysis}
\subsection{A dense index lowers R@1}
\label{sec:drop}
We evaluate Sample4Geo~\cite{deuser2023sample4geo} on both benchmarks, together with one recent method per benchmark whose released weights or code reproduce its published R@1: the retrieval stage of UnifyGeo~\cite{song2025unified} on VIGOR and AuxGeo~\cite{xia2025cross} on DReSS. Sample4Geo and AuxGeo have no released DReSS weights, so we retrain them to within 1.1 percentage points of the published R@1 (\cref{sec:supp_repro}). We additionally train our own whole-tile model, a DINOv3 ViT-B~\cite{simeoni2025dinov3} with the loss and sampling strategy of Sample4Geo (\cref{sec:setup}). It shares backbone, initialization and schedule with LaSo, whereas the published models use other backbones, such as the ConvNeXt-B of Sample4Geo and AuxGeo. A difference between LaSo and this model thus comes from the positive and its siblings, not from the architecture or its weights.

Every model loses R@1 when the dense index replaces the standard gallery, although the index still contains every tile (\cref{tab:dense_drop}): 6.9--12.8 percentage points on VIGOR and 0.4--6.5 on DReSS. On DReSS, the loss is small for Sample4Geo and our model, at most 1.6 points.

The loss does not come from retrievals far from the camera. The median localization error falls for every model (\cref{tab:dense_drop}), and R@5 rises (\cref{tab:supp_r5}). For every model and split, the retrieved tile is one of the eight neighboring tiles for 10.0--17.7 percentage points more of the queries than on the standard gallery, and lies four or more grid steps from the positive for 2.0--11.0 points fewer (\cref{tab:supp_errtypes}). On DReSS, the second effect offsets most of the first, which keeps the loss small there. The dense index thus brings the best crop closer to the camera, yet this crop more often belongs to a neighboring tile.
\subsection{Whole-tile training leads to a plateau}
\label{sec:plateau}
\paragraph{Mean similarity and residual.} To see what a model has learned about position, we split the similarity of a query and a reference into three parts,
\begin{equation}
    s(q, r) = a_q + \bar s(\boldsymbol{\delta}_r) + e_{q,r},
    \label{eq:score_model}
\end{equation}
and fit the level $a_q$ of each query and the \emph{mean similarity} $\bar s$, which has one value per bin of the offset, by least squares (\cref{sec:supp_simabl}).
The level is shared by all references of a query and does not affect their ranking. The fit makes $\bar s(\boldsymbol{\delta})$ the average of $s(q, r) - a_q$ over the pairs at offset $\boldsymbol{\delta}$ and leaves the \emph{residual} $e_{q,r}$ with zero mean at every offset. The residual has standard deviation $\sigma$ and reflects the image content of the individual reference. The split is an identity and holds for any model, since the residual is whatever $a_q$ and $\bar s$ leave. What training determines is the shape of $\bar s$: where $\bar s$ differs less between two references than their residuals do, image content rather than distance to the camera decides which of them ranks higher.
\paragraph{What the training objective asks for.} A whole-tile model is trained with the positive as the only positive of a query and with the other tiles in the batch as negatives, semi-positives included. Seen from the camera, the center of the positive lies in $\mathcal{S}$, the square of side $g$ around the camera, at every position about equally often (\cref{sec:setting}), and the center of every negative lies outside $\mathcal{S}$. The objective therefore asks for a high similarity everywhere in $\mathcal{S}$ and a low one everywhere else, and it is indifferent to the order of two references inside $\mathcal{S}$. Among mean similarities that depend on the offset alone, it is idealized by a step function,
\begin{equation}
    \bar s^\ast(\boldsymbol{\delta}) =
    \begin{cases}
        s_{\mathrm{hi}}, & \|\boldsymbol{\delta}\|_\infty \le g/2, \\
        s_{\mathrm{lo}}, & \text{otherwise},
    \end{cases}
    \label{eq:plateau}
\end{equation}
where $s_{\mathrm{hi}}$ and $s_{\mathrm{lo}}$ are the highest and lowest mean similarities the model can attain. We call the region of constant mean similarity around the camera a \emph{plateau}. In this ideal case it is the square $\mathcal{S}$, and a reference at distance $0.4g$ from the camera has the same mean similarity as one at the camera. A trained model can only approximate it.
\begin{figure}[t]
    \centering
    \includegraphics[width=0.95\linewidth]{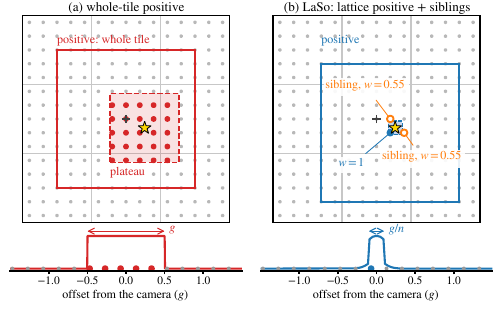}
    \caption{\textbf{Training positives and the mean similarity they lead to}, schematic for $n = 5$. Dots: crop centers. Gray lines: cell borders. $+$: tile center. Star: camera. (a)~Whole tile: the ideal plateau is the square $\mathcal{S}$ (shaded), which holds $n^2$ crop centers, some of them in neighboring cells. (b)~LaSo: the positive is the crop nearest to the camera (square: its footprint), and the plateau shrinks to side $g/n$. The two siblings are lattice neighbors of the positive in the same tile, weighted by \cref{eq:sib_weights}. Bottom: the training target along a line through the camera, averaged over the camera positions in the cell.}
    \label{fig:method}
\end{figure}
\paragraph{Consequence for the two galleries.} The standard gallery places exactly one tile center in $\mathcal{S}$, that of the positive, so the plateau costs nothing there. The dense index places $n^2-1$ additional crop centers in $\mathcal{S}$. The square $\mathcal{S}$ is centered on the camera, whereas the crops that belong to the positive lie in its cell, which is centered on the tile. The two squares coincide only for a camera at the tile center, so some of the $n^2$ crops belong to neighboring tiles (\cref{fig:method}a). On the ideal plateau, if the residuals favor none of the crops, each of them is the best crop equally often. For a camera uniformly distributed in its cell, the best crop then belongs to a neighboring tile with probability $1 - \big(\tfrac34 + \tfrac{1}{4n^2}\big)^2$, which is zero for $n = 1$ and 42\% for $n = 5$ (\cref{sec:supp_share}). We call such a retrieval a \emph{neighbor miss}.
\paragraph{Measurement.} We measure $\bar s$ in 14 cases: Sample4Geo on all four splits, UnifyGeo on both VIGOR splits, AuxGeo on both DReSS splits, and six of our whole-tile ViT-B models (\cref{sec:supp_simabl}). Each case uses 5\,000 test queries and, for each query, the crops of the positive and its eight neighboring tiles. We make three observations.

\noindent(i)~\textbf{The mean similarity is nearly flat around the camera.} Within $0.3g$ of the camera, $\bar s$ falls by only 0.01--0.03 (\cref{fig:score_field}). The ranking reflects this. The crop nearest to the camera has a higher similarity than a crop $0.2$--$0.3g$ away in only 57--70\% of the cases. For LaSo, it is 97--98\%.

\noindent(ii)~\textbf{Neighbor misses are about as frequent as on the ideal plateau.} The best of these crops belongs to a neighboring tile for 31--46\% of the queries (38\% on average), against 42\% for the ideal plateau.

\noindent(iii)~\textbf{The plateau follows the positive.} Nearby crops share most of their imagery, which could flatten $\bar s$ on its own. We therefore change only the positive and fine-tune one ViT-B with positives centered on the camera plus an offset drawn uniformly from $[-J, J]^2$, from $J = 0$ to $J = 0.625g$, beyond the whole-tile case $J = g/2$. The radius at which $\bar s$ has fallen by 0.05 grows linearly with $J$, on VIGOR from $0.145g$ at $J = 0$ to $0.46g$ at $J = g/2$, and the whole-tile models lie close to the same line. This holds for every threshold we tried (correlation above 0.97, \cref{sec:supp_simabl}), and R@1 on the $5\times5$ index falls as the radius grows, on VIGOR from 86.6 to 52.2 (\cref{tab:supp_jmiss}). The plateau thus comes from the training positive.
\subsection{Reading out the plateau: a spatial vote}
\label{sec:vote}
On a plateau, the single best crop is unreliable, but the plateau is centered on the camera, so the crops most similar to a query cluster around the camera. Where they are densest is a better position estimate than any single crop. Let $\mathcal{T}_q$ be the 200 crops most similar to query $q$ and $s_{\max}$ the highest similarity among them. Each crop $r' \in \mathcal{T}_q$ votes for the crops near it, with a weight $\omega_{r'} = \exp\!\big((s(q, r') - s_{\max})/\kappa\big)$ that decays with its similarity gap to the best crop:
\begin{equation}
    V(r) = \sum_{r' \in \mathcal{T}_q} \omega_{r'} \exp\!\Big(-\frac{\|\mathbf{c}_{r'} - \mathbf{c}_r\|_2^2}{2b^2}\Big), \qquad r \in \mathcal{T}_q.
    \label{eq:vote}
\end{equation}
This is a similarity-weighted kernel density estimate of where the most similar crops lie, as used in retrieval-based geolocation~\cite{vo2017revisiting}. The crop with the largest $V$ replaces the best crop, and we call a model read out in this way \emph{voted}.

Neither benchmark has a validation split, so we select the readout rule and its parameters on one benchmark and apply them to the other. Both directions select $\kappa = 0.1$ and $b$ equal to two strides (\cref{sec:supp_rules}). With 50 to 1\,000 crops in $\mathcal{T}_q$ instead of 200, R@1 changes by at most 0.8 points.

With the vote, every whole-tile model exceeds its own standard-gallery R@1, by 2.5--6.4 points on VIGOR and 7.1--10.8 on DReSS (\cref{tab:dense_drop}), and its median localization error is roughly half that without the vote. The vote changes how the similarities are read, not the similarities themselves, so shrinking the plateau requires a different training positive.

\section{LaSo: Matching the Positive to the Index}
\label{sec:method}
\Cref{sec:plateau} showed that the plateau follows the square in which the center of the training positive can lie. LaSo therefore shrinks this square, and the plateau with it (\cref{fig:score_field}). It takes the lattice crop nearest to the camera as the positive, which leaves a square of one stride, and adds neighboring crops as soft targets weighted by their distance to the camera (\cref{fig:method}b).

\subsection{Lattice positive}
\label{sec:lattice_pos}
For a training pair with camera position $\mathbf{x}$ in the cell of tile $t$, LaSo takes the crop of $t$ whose center is nearest to the camera as the positive,
\begin{equation}
    r^\ast = \operatorname*{arg\,min}_{r} \|\mathbf{c}_r - \mathbf{x}\|_2,
    \label{eq:lattice_pos}
\end{equation}
where $r$ runs over the $n^2$ crops of $t$, cut from the mosaic of the training tiles as in \cref{eq:lattice}. The center of this positive lies within half a stride of the camera along each axis, $\|\boldsymbol{\delta}_{r^\ast}\|_\infty \le g/(2n)$, \ie, in a square of side $g/n$ around the camera instead of in $\mathcal{S}$. This is the case $J = g/(2n)$ of \cref{sec:plateau}, with $J$ set by the lattice of the index instead of chosen as a hyperparameter. The ideal plateau shrinks to this square, which contains exactly one crop of the index, the one nearest to the camera.

A positive centered exactly on the camera, $J = 0$, may seem more direct. The nearest crop of the index, however, lies up to $0.14g$ from the camera, half a stride per axis. A positive with $J = 0$ gives a peak whose radius (\cref{sec:plateau}) is $0.145g$ on VIGOR but $0.071g$ on DReSS, so on DReSS the nearest crop can fall on its slope, which costs 13.4 points of R@1 against the lattice positive (\cref{sec:ablation}).
\subsection{Soft siblings}
\label{sec:siblings}
With the lattice positive alone, each training pair contributes one crop to the batch. A query is then never compared with the other crops of its own tile, although these are among its nearest competitors in the index. LaSo therefore adds $M = 2$ of them to each pair; more siblings change R@1 on the $5\times5$ index by at most 0.7 points (\cref{sec:supp_siblings}). These \emph{siblings} are drawn without replacement from the up to eight lattice neighbors of the positive within the same tile. Treating them as negatives would ask the model to separate crops that can be almost as close to the camera as the positive. Instead, each sibling is a soft target whose weight decreases with its distance to the camera. Let $d_{i,0}$ be the distance from the camera of pair $i$ to the center of its positive and $d_{i,m}$, $m = 1, \dots, M$, the distance to sibling $m$. The weight is
\begin{equation}
    w_{i,m} = \exp\!\Big(-\frac{d_{i,m}^2 - d_{i,0}^2}{2\gamma^2}\Big), \qquad \gamma = \frac{g}{2n},
    \label{eq:sib_weights}
\end{equation}
so that the positive has $w_{i,0} = 1$ and a sibling as close to the camera as the positive receives the same weight. We set the width $\gamma$ to half a stride, the half-width of the square in which the center of the positive lies, and do not tune it.

\subsection{Loss}
\label{sec:loss}
Let $\ell_{i,j} = s(q_i, r_j)/\tau$ be the logit of query $q_i$ and reference $r_j$, with the learned temperature $\tau$. A batch of $B$ pairs contains $B(1 + M)$ references, the positive and the $M$ siblings of each pair. For query $q_i$, the target of reference $r_j$ is
\begin{equation}
    y_{i,j} = (1 - \varepsilon)\, \bar w_{i,j} + \frac{\varepsilon}{B(1 + M)},
    \label{eq:sib_target}
\end{equation}
where $\bar w_{i,j} = w_{i,m} / \sum_{m'} w_{i,m'}$ if $r_j$ is the positive ($m = 0$) or sibling $m$ of pair $i$, and zero otherwise, and $\varepsilon = 0.1$ is the label smoothing of Sample4Geo~\cite{deuser2023sample4geo}. The loss is the cross-entropy between these targets and the softmax of the logits over all references, $\mathcal{L} = -\frac{1}{B}\sum_{i}\sum_{j} y_{i,j} \log\big(e^{\ell_{i,j}} / \sum_{k} e^{\ell_{i,k}}\big)$. The positives and siblings of the other pairs thus remain negatives. As in Sample4Geo, we add the reference-to-query direction with equal weight, over the $B$ positives only, since siblings have no query of their own.

The cross-entropy is minimal where the softmax equals the targets, \ie, up to the label smoothing, where the logits of the positive and of sibling $m$ differ by $\log(w_{i,0}/w_{i,m}) = (d_{i,m}^2 - d_{i,0}^2)/(2\gamma^2)$. The loss thus asks for a similarity that falls with the squared distance to the camera, a peak in place of the plateau.

\subsection{Inference}
\label{sec:inference}
LaSo is evaluated with the protocol of \cref{sec:protocol}, on an index with the lattice it was trained for. The best crop gives the retrieved tile and a first position estimate. We refine the position by a soft-argmax. Among the 16 crops most similar to the query, we keep the set $\mathcal{N}$ of those within 1.5 strides of the best crop and estimate the camera position as $\hat{\mathbf{x}} = \sum_{r \in \mathcal{N}} \pi_r \mathbf{c}_r$ with $\pi_r \propto \exp(s(q, r)/\tau)$. Neither number is tuned; with 9 to 64 crops instead of 16, the median error changes by at most 0.07\,m. The retrieved tile does not change.

\section{Experiments}
\label{sec:experiments}

\subsection{Setup}
\label{sec:setup}
\paragraph{Implementation.} We use a shared DINOv3 backbone~\cite{simeoni2025dinov3} for both ground-level and overhead images, ViT-B/16 or ViT-L/16, whose patch tokens we average into the embedding. Overhead images are resized to $384\times384$ pixels and panoramas to $384\times768$. All models are trained using AdamW~\cite{loshchilov2019decoupled}, a learning rate of $5\cdot10^{-5}$, a cosine learning-rate schedule, and batches of 128 pairs built from geographic and visual neighbors as in Sample4Geo~\cite{deuser2023sample4geo} (\cref{sec:supp_training}). We report the last epoch of every run. For every LaSo model, we train a whole-tile model with the same backbone, initialization, seed, and schedule. It uses the tile as the positive and no siblings, which reduces the loss of \cref{sec:loss} to that of Sample4Geo.

\paragraph{Initialization.} Recent gains in cross-view retrieval come largely from more pre-training data~\cite{huang2024cv,geobridge2026,geobridgepp2025,zhang2026geo}, so we test whether such data closes the gap between whole-tile training and LaSo. The \emph{public} initialization starts from the released DINOv3 weights (LVD-1689M) and trains on the benchmark for 40 epochs. \emph{CC pre-training} first trains for 30 epochs on 202\,k camera-centered pairs from CVUSA~\cite{workman2015wide}, CVACT~\cite{liu2019lending}, and CV-Cities~\cite{huang2024cv}, none within 50\,km of a VIGOR or DReSS test city, and then for 20 epochs on the benchmark. The ablations fine-tune the ViT-B with CC pre-training for five epochs. We use three seeds for every ViT-B model and for LaSo ViT-L with CC pre-training, and one otherwise.

\begin{table}[t]
    \centering
    \footnotesize
    \setlength{\tabcolsep}{3.2pt}
    \begin{tabular}{@{}llcccc@{}}
        \toprule
        & & \multicolumn{2}{c}{VIGOR} & \multicolumn{2}{c}{DReSS} \\
        \cmidrule(lr){3-4}\cmidrule(lr){5-6}
        Method & Index & cross & same & cross & same \\
        \midrule
        \multicolumn{6}{@{}l}{\emph{Published}} \\
        Sample4Geo~\cite{deuser2023sample4geo} & Tiles & 61.70 & 77.86 & 28.23$^\ast$ & 51.40$^\ast$ \\
        AuxGeo~\cite{xia2025cross} & Tiles & 63.94 & 80.34 & \textbf{32.44} & \textbf{54.70} \\
        UnifyGeo$^\ddagger$~\cite{song2025unified} & Tiles & 67.58 & 82.80 & -- & -- \\
        GeoBridge++$^\dagger$~\cite{geobridgepp2025} & Tiles & 78.20 & \textbf{89.49} & -- & -- \\
        C-BEV$^\ddagger$~\cite{fervers2023c} & Tiles & \textbf{80.01} & 87.49 & -- & -- \\
        \midrule
        \multicolumn{6}{@{}l}{\emph{ViT-B, public initialization}} \\
        Whole tile & Tiles & 64.32 & 76.96 & 33.37 & 51.03 \\
        Whole tile & Dense +Vote & 70.75 & 80.65 & 42.99 & 61.83 \\
        LaSo & Dense & \textbf{85.98} & \textbf{89.95} & \textbf{76.31} & \textbf{87.49} \\
        \midrule
        \multicolumn{6}{@{}l}{\emph{ViT-L, public initialization}} \\
        Whole tile & Tiles & 72.99 & 81.35 & 46.93 & 64.00 \\
        Whole tile & Dense +Vote & 77.64 & 83.88 & 56.46 & 72.40 \\
        LaSo & Dense & \textbf{89.99} & \textbf{92.21} & \textbf{85.17} & \textbf{92.39} \\
        \midrule
        \multicolumn{6}{@{}l}{\emph{ViT-B, CC pre-training}} \\
        Whole tile & Tiles & 71.67 & 78.64 & 43.21 & 55.57 \\
        Whole tile & Dense +Vote & 76.60 & 81.91 & 54.47 & 66.66 \\
        LaSo & Dense & \textbf{88.09} & \textbf{90.01} & \textbf{80.18} & \textbf{87.80} \\
        \midrule
        \multicolumn{6}{@{}l}{\emph{ViT-L, CC pre-training}} \\
        Whole tile & Tiles & 77.98 & 82.65 & 55.70 & 66.97 \\
        Whole tile & Dense +Vote & 81.26 & 84.53 & 65.55 & 74.86 \\
        LaSo & Dense & \textbf{90.66} & \textbf{92.03} & \textbf{86.85} & \textbf{92.30} \\
        \bottomrule
    \end{tabular}
    \caption{\textbf{R@1 of published methods and of LaSo} on the standard gallery (Tiles) and on the $5 \times 5$ dense index with the best crop (Dense) or the spatial vote (Dense +Vote). Published: as reported, $^\ast$by AuxGeo, $^\dagger$with extra training data, $^\ddagger$with a re-ranking stage. Ours: mean over three seeds for ViT-B and for LaSo ViT-L with CC pre-training, one seed otherwise, with a standard deviation of at most 0.7 points. LaSo is read out with the best crop, as the vote lowers its R@1 by 0.1--3.6 points. \Cref{tab:supp_full,tab:supp_published} list all readouts and more methods. Best values per block in bold.}
    \label{tab:main}
\end{table}
\begin{table}[t]
    \centering
    \small
    \setlength{\tabcolsep}{3pt}
    \resizebox{\columnwidth}{!}{%
    \begin{tabular}{@{}l cc cc cc@{}}
        \toprule
        & \multicolumn{2}{c}{Median [m]} & \multicolumn{2}{c}{Within 1\,m [\%]} & \multicolumn{2}{c}{Within 10\,m [\%]} \\
        \cmidrule(lr){2-3}\cmidrule(lr){4-5}\cmidrule(lr){6-7}
        Method & cross & same & cross & same & cross & same \\
        \midrule
        \multicolumn{7}{@{}l}{\emph{Correct tile given, queries in its cell}} \\
        CCVPE~\cite{xia2023convolutional} & 1.68 & 1.36 & -- & -- & -- & -- \\
        HC-Net~\cite{wang2023fine} & 1.59 & 1.17 & -- & -- & -- & -- \\
        FG$^2$~\cite{xia2025fg} & \textbf{1.37} & \textbf{1.08} & -- & -- & -- & -- \\
        \midrule
        \multicolumn{7}{@{}l}{\emph{Whole test area, all queries}} \\
        Sample4Geo + CCVPE~\cite{song2025unified} & -- & -- & 18.61 & 29.90 & 56.71 & 73.10 \\
        Sample4Geo + HC-Net~\cite{song2025unified} & -- & -- & 19.28 & 35.92 & 59.23 & 74.91 \\
        UnifyGeo~\cite{song2025unified} & -- & -- & 25.58 & \textbf{39.64} & 66.24 & 81.37 \\
        Whole tile ViT-B + vote & 5.56 & 4.92 & 2.25 & 2.91 & 70.46 & 77.21 \\
        Whole tile ViT-L + vote & 5.06 & 4.62 & 2.57 & 3.12 & 76.31 & 80.86 \\
        LaSo ViT-B & 1.72 & 1.48 & 22.27 & 29.84 & 87.49 & 89.46 \\
        LaSo ViT-L & \textbf{1.52} & \textbf{1.29} & \textbf{27.16} & 36.08 & \textbf{90.21} & \textbf{91.60} \\
        \bottomrule
    \end{tabular}}
    \caption{\textbf{Metric localization on VIGOR}, cross and same area. Top: fine-grained methods with known orientation, given the correct tile and the queries in its cell, as reported. CCVPE and FG$^2$ measure against the corrected labels of SliceMatch~\cite{lentsch2023slicematch}, which differ from the GPS positions by 0.1\,m on average, HC-Net against its own. Bottom: retrieval over the whole test area and all queries. Within 1\,m and 10\,m: the retrieved tile is the positive and the error is at most 1\,m or 10\,m, as in the UnifyGeo code, published pipelines as reported by UnifyGeo. Ours: CC pre-training, errors to the GPS position, whole tile at the crop of the spatial vote, LaSo at the soft-argmax, mean over three seeds (whole-tile ViT-L: one). Best values per block in bold.}
    \label{tab:metric}
\end{table}

\subsection{Main results}
\label{sec:main}
\paragraph{Retrieval.} LaSo reaches the highest R@1 in \cref{tab:main} on every split, with both backbones and both initializations. It exceeds the voted whole-tile model with the same backbone and initialization by 7.5--33.3 percentage points, least on VIGOR same-area and most on DReSS cross-area. LaSo removes most neighbor misses: its best crop belongs to a neighboring tile for 4.7--7.3\% of the queries, against 17.8--30.4\% for the whole-tile models. On VIGOR, this is most of its gain. The plateau accounts for neighbor misses only. On DReSS, LaSo also cuts the misses four or more grid steps away to about a third (\cref{tab:supp_errtypes}). These follow the training positive as well: in the ablation of \cref{sec:ablation}, their share on DReSS cross is 31\% for the whole tile, 12--13\% for the lattice positive and for $J = 0.075g$ to $0.125g$, with or without siblings, and 24--39\% for all other $J$ (\cref{tab:supp_jmiss}). A positive matched to the index thus also separates distant tiles better, and this does not depend on the dense index: on the standard gallery, for the DReSS queries nearest to the tile center (subset S1 of~\cite{xia2025cross}), LaSo reaches 89.0\% on cross-area against 59.6\% for the whole-tile model (\cref{tab:decentrality}). On the dense index, LaSo stays within one point from S1 to S3 and loses 9--10 points in S4, to neighbor misses at the border of the cell. A stronger whole-tile model does not close the gap on its own. From ViT-B with public initialization to ViT-L with CC pre-training, its loss on the dense index grows from 6.9 to 13.1 points on VIGOR cross-area (\cref{tab:supp_full}), and CC pre-training narrows the gap to LaSo without closing it. With the ConvNeXt-B~\cite{liu2022convnet} of Sample4Geo and AuxGeo, LaSo reaches 81.64 and 69.43 on VIGOR and DReSS cross-area (\cref{tab:supp_published}), against 66.06 and 39.38 for their voted models.

\paragraph{Localization.} The best crop of LaSo has a median localization error of 3.2--3.4\,m on VIGOR and 7.6--9.4\,m on DReSS, against 4.6--6.3\,m and 13.8--39.4\,m for our voted whole-tile models. The soft-argmax of \cref{sec:inference} roughly halves it, to 1.3--3.5\,m for LaSo ViT-L with CC pre-training. \Cref{tab:metric} compares on VIGOR. Fine-grained methods~\cite{xia2023convolutional,wang2023fine,xia2025fg}, which are given the positive and evaluate only the queries in its cell, reach 1.4--1.7\,m on cross-area. LaSo reaches 1.52\,m while retrieving over the whole test area and counting all queries, misses included. In the protocol of UnifyGeo~\cite{song2025unified}, it localizes 90.2\% of the cross-area queries within 10\,m, against 66.2\% for UnifyGeo, and 27.2\% against 25.6\% within 1\,m. On same-area, UnifyGeo is ahead within 1\,m, with 39.6\% against 36.1\%.

\begin{table}[t]
    \centering
    \footnotesize
    \setlength{\tabcolsep}{2.6pt}
    \begin{tabular}{@{}l ccc ccc@{}}
        \toprule
        & \multicolumn{3}{c}{VIGOR cross} & \multicolumn{3}{c}{DReSS cross} \\
        \cmidrule(lr){2-4}\cmidrule(lr){5-7}
        Positive & $n{=}1$ & $n{=}3$ & $n{=}5$ & $n{=}1$ & $n{=}3$ & $n{=}5$ \\
        \midrule
        Whole tile & 69.0 & 63.7 & 63.0 & 38.6 & 39.1 & 39.3 \\
        Camera, $J = 0$ & 30.6 & 83.5 & 86.6 & 8.0 & 36.9 & 61.9 \\
        Camera, $J = 0.075g$ & 36.8 & 84.7 & 85.8 & 15.4 & 62.2 & 75.9 \\
        Camera, $J = 0.125g$ & 45.2 & 84.4 & 83.8 & 22.3 & 69.1 & 73.7 \\
        Camera, $0.075g$, 2 adjacent, soft & 30.0 & 84.5 & 87.6 & 18.0 & 67.7 & 77.7 \\
        Lattice, no siblings & 39.8 & 84.7 & 84.9 & 18.5 & 66.9 & 75.4 \\
        Lattice, 2 of other tiles & 36.5 & 85.2 & 86.5 & 17.9 & 66.6 & 76.6 \\
        Lattice, 2 random, hard & 24.4 & 81.9 & 87.1 & 13.2 & 59.9 & 76.9 \\
        Lattice, 2 random, soft & 30.5 & 84.8 & 87.3 & 17.7 & 67.9 & 77.5 \\
        LaSo: 2 adjacent, soft & 31.9 & 85.0 & 87.4 & 18.7 & 68.4 & 77.2 \\
        \bottomrule
    \end{tabular}
    \caption{\textbf{R@1 for different training positives} on indexes with $n \times n$ crops per tile, $n = 1$ being the standard gallery. Camera: centered on the camera plus an offset uniform in $[-J, J]^2$. Of other tiles: two more crops per pair, as negatives. Hard siblings are negatives, soft ones weighted by \cref{eq:sib_weights}. Random siblings come from the whole tile, adjacent ones from around the positive. Five-epoch fine-tunes of the ViT-B with CC pre-training, mean over three seeds, standard deviation at most 0.7 points, see \cref{tab:supp_ablation}.}
    \label{tab:ablation}
\end{table}
\subsection{Ablation of the positive}
\label{sec:ablation}
\Cref{tab:ablation} varies the training positive and its siblings, using the fine-tunes of observation (iii) in \cref{sec:plateau}.

The optimal positive depends on the index. Among these positives, the whole tile performs best on the standard gallery ($n=1$), whereas on the $5\times5$ index, all tested positives near the camera outperform it by 21--38 points.

The lattice positive is a camera-centered positive with $J = g/(2n) = 0.1g$, restricted to the crops of the index, and its R@1 lies between those of $J = 0.075g$ and $J = 0.125g$ in every column. What it removes is the search for $J$, whose best value depends on the benchmark and on $n$. With $J = 0$, the positive is 1.7 points ahead of the lattice positive on VIGOR at $n=5$ but 13.4 points behind on DReSS, where the nearest crop can lie outside the narrow peak that this positive learns (\cref{sec:lattice_pos}). Without a search, the lattice positive stays within 1.7 points of the test-selected best $J$ on the $5\times5$ index.

Two siblings add 2.5 points on VIGOR and 1.9 on DReSS at $n=5$. Most of this comes from the larger number of references in the batch: two crops of other tiles per pair add 1.6 and 1.2 points. The siblings account for the remaining 0.9 and 0.6 points and lower the median error of the soft-argmax from 2.0 to 1.8\,m on VIGOR and from 5.2 to 4.7\,m on DReSS (\cref{sec:supp_siblings}). With the same two soft siblings, a camera-centered positive at $J = 0.075g$, the best $J$ on DReSS, is 0.2 and 0.5 points ahead of LaSo at $n = 5$ and 0.5--1.9 behind at $n = 1$ and 3. Both positives match the stride of the index, as the analysis requires, and the lattice does so without a search, at a cost of at most 0.5 points. Soft and hard siblings perform similarly at $n=5$, but soft targets are ahead on coarser indexes, by up to 8.0 points at $n=3$ and 6.0 points at $n=1$. Random and adjacent siblings are within seed variation at $n = 5$; we keep the adjacent ones, the crops that compete with the best crop on the index.

A finer lattice helps, with diminishing returns. With matched training and index, R@1 rises by 2.5 points on VIGOR and 7.7 on DReSS from $n = 3$ to 5, and by 0.5 and 2.8 from 5 to 7, while the index grows from 9 to 25 to 49 crops per tile (\cref{sec:supp_lattice}). On DReSS, the model trained with $n = 5$ gains half of the last step from the $7\times7$ index alone. We use $n = 5$ as a compromise between accuracy and index size.
\section{Limitations and Conclusion}
\paragraph{Limitations.} LaSo relies on a dense index that stores 25 embeddings per tile and thus needs 25 times the storage of a standard gallery. The index is embedded once, a query still takes one forward pass, and only the search grows, by the same factor. Compressing the embeddings reduces this storage with almost no loss of R@1, and a $3\times3$ lattice reduces it to nine times at a cost of up to 3.3 percentage points on VIGOR and 8.2 on DReSS (\cref{sec:supp_cost,tab:supp_n3}). LaSo is designed for the dense index and does not transfer directly to the standard gallery, where its R@1 drops to 18--35\% because the original tile centers generally lie outside the narrow similarity peak learned during training. LaSo needs training camera positions that are accurate to within a stride, whereas whole-tile training only needs the tile. Like the compared methods, LaSo assumes north-aligned panoramas, and unknown orientation is left for future work.

\paragraph{Conclusion.} Dense reference sampling has been reported not to help cross-view retrieval. We traced this to the training positive: a whole tile may lie anywhere within a grid step of the camera, the learned similarity forms a plateau of that size, and on a dense index image content, not distance, decides among the crops on it. To our knowledge, this is the first analysis of how the training positive shapes the similarity around the camera. A spatial vote reads the plateau out and lifts existing models above their standard-gallery R@1 without retraining.

Matching the positive to the index shrinks the plateau to one stride. Changing only the positive and its targets, LaSo turns the dense index into the better gallery: it exceeds whole-tile training of the same backbone on the standard gallery by 9.4--42.9 points of R@1, on every split, backbone and initialization. With ViT-L, its median error of 1.5\,m on VIGOR cross-area lies in the range of fine-grained methods that are given the correct tile, with one embedding per crop and no second model. This suggests that retrieval and fine-grained localization can be served by one model once the training positive reflects the index.

\section*{LLM Usage Statement}
We used LLM assistants (Claude, Claude Code, Codex) to edit the manuscript, to help create figures, and to write code for training and evaluation. No labels or data were LLM-generated, and the authors verified all results and take full responsibility for the content.

{
    \small
    \bibliographystyle{ieeenat_fullname}
    \bibliography{main}
}

\clearpage
\setcounter{page}{1}
\makeatletter\let\Hy@backout\@gobble\makeatother
\maketitlesupplementary
\setcounter{section}{0}\setcounter{figure}{0}\setcounter{table}{0}\setcounter{equation}{0}
\renewcommand{\thesection}{S\arabic{section}}\renewcommand{\thefigure}{S\arabic{figure}}\renewcommand{\thetable}{S\arabic{table}}\renewcommand{\theequation}{S\arabic{equation}}
\makeatletter\renewcommand{\theHsection}{S\arabic{section}}\renewcommand{\theHfigure}{S\arabic{figure}}\renewcommand{\theHtable}{S\arabic{table}}\renewcommand{\theHequation}{S\arabic{equation}}\makeatother

\section{Camera positions within the cell}
\label{sec:supp_cell}
\begin{figure}[t]
    \centering
    \includegraphics[width=\linewidth]{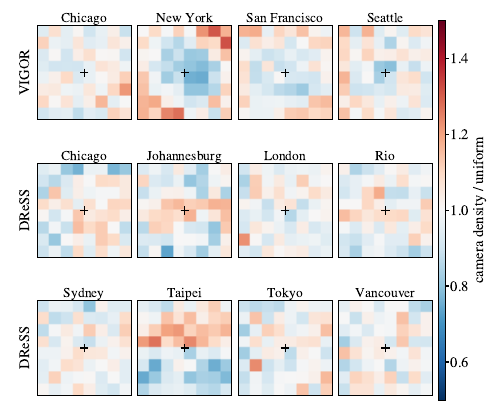}
    \caption{\textbf{Where the test cameras lie within the cell of their positive tile.} Density of the camera positions relative to a uniform distribution, per city, on an $8\times8$ grid over the cell (+: tile center, north up). White means uniform. Test queries of the same-area splits, which cover every city, 10--14\,k per city.}
    \label{fig:supp_cell}
\end{figure}
On VIGOR, the grid step is $g = 36.5$\,m in all four cities, while the ground resolution of the 640-pixel tiles differs per city, from 0.101\,m per pixel in Seattle to 0.118\,m in San Francisco~\cite{lentsch2023slicematch}. A tile thus covers 64--76\,m, and neighboring tiles overlap by 43--52\%. VIGOR's original camera positions assume one resolution for all cities and are off by up to 3\,m~\cite{lentsch2023slicematch}, so we take the camera positions from GPS. On DReSS, $g = 73$--$108$\,m and a tile covers 83--123\,m, depending on the city's latitude, so neighboring tiles overlap by 12.5\%. Several of our arguments assume that a test camera can lie anywhere in the cell of its positive tile. \Cref{fig:supp_cell} shows that this holds in every city of both benchmarks. The share of cameras in the inner square of half the cell's side, where both coordinates lie within a quarter grid step of the tile center, is 21.0--26.3\% against 25\% for a uniform distribution, and the density stays within 0.74--1.32 times the uniform value in every bin. On DReSS, about 1\% of the cameras lie just outside the cell of their labeled positive. They sit on the border between two cells, 90\% of them within 0.5\% of a grid step, and the dataset labels the neighboring tile, whose center is almost equally near, as their positive. For the other 99.0\% the labeled positive is the tile with the nearest center. On VIGOR, this holds for 99.97\% of the test cameras.

\section{Reproduction of the compared models}
\label{sec:supp_repro}
\Cref{tab:dense_drop} includes Sample4Geo on both benchmarks and one recent method per benchmark whose released weights or code reproduce its published R@1. \Cref{tab:supp_repro} compares their standard-gallery R@1 with the published numbers. The released weights of GeoBridge~\cite{geobridge2026} are pre-trained on its multi-view data but not fine-tuned on VIGOR, so we do not evaluate it.

\paragraph{Released checkpoints.} For Sample4Geo~\cite{deuser2023sample4geo} on VIGOR, we use the released ConvNeXt-B checkpoints of the last training epoch. Their R@1, as stored with the released weights, is 61.31 / 77.82 on VIGOR cross / same, slightly below the 61.70 / 77.86 of the paper. For UnifyGeo~\cite{song2025unified}, we use the released checkpoints of the retrieval stage, whose global descriptor is what an index stores, and embed queries and references with UnifyGeo's own preprocessing. It matches the published R@1 of the retrieval stage to within 0.07 points.

\paragraph{Training on DReSS.} Neither AuxGeo~\cite{xia2025cross} nor Sample4Geo releases weights for DReSS, so we trained both with their official code. We changed only the data paths and, for Sample4Geo, the data loader, which reads the DReSS cities and the GPS neighbor lists shipped with DReSS instead of VIGOR's. Both runs follow the recipe of their code: 40 epochs, a ConvNeXt-B initialized from ImageNet-22k with ImageNet-1k fine-tuning (the 224-pixel checkpoint for AuxGeo and the 384-pixel one for Sample4Geo, as in the respective code), a batch of 128 pairs, AdamW with a learning rate of $10^{-3}$, a cosine schedule with one warm-up epoch, label smoothing of 0.1, and the hard-negative sampling of Sample4Geo, which draws neighbors by GPS for the first four epochs and by embedding similarity afterwards, refreshed every four epochs. We distribute each batch over eight GPUs, which leaves the gradient of the full batch unchanged, and report the last epoch. Two points deviate from the AuxGeo code. Its DReSS training script sets a batch of 12, whereas the paper states 128, which we use. The code keeps the checkpoint with the best test R@1, whereas we report the last epoch, which may account for our 0.2--0.4 points lower R@1. Sample4Geo reaches 29.25 / 51.98 on DReSS cross / same, 1.0 / 0.6 points above the numbers that AuxGeo reports for it.

\paragraph{Evaluation.} All numbers in \cref{tab:dense_drop} come from our evaluation, which embeds tiles and crops with each model's own network and preprocessing. On the standard gallery, it agrees with the official evaluation scripts to within 0.07 points of R@1, see \cref{tab:supp_repro}. The remaining differences are numerical.

\begin{table}[h]
    \centering
    \small
    \setlength{\tabcolsep}{3pt}
    \begin{tabular}{@{}llccc@{}}
        \toprule
        Model & Split & Published & Official & Ours \\
        \midrule
        \multicolumn{5}{@{}l}{\emph{Released checkpoints}} \\
        Sample4Geo~\cite{deuser2023sample4geo} & VIGOR cross & 61.70 & 61.31 & 61.38 \\
                                               & VIGOR same  & 77.86 & 77.82 & 77.88 \\
        UnifyGeo~\cite{song2025unified}        & VIGOR cross & 61.18 & --    & 61.25 \\
                                               & VIGOR same  & 78.44 & --    & 78.47 \\
        \midrule
        \multicolumn{5}{@{}l}{\emph{Trained with the official code}} \\
        AuxGeo~\cite{xia2025cross}             & DReSS cross & 32.44 & 32.21 & 32.27 \\
                                               & DReSS same  & 54.70 & 54.32 & 54.37 \\
        Sample4Geo~\cite{deuser2023sample4geo} & DReSS cross & 28.23$^\dagger$ & 29.25 & 29.29 \\
                                               & DReSS same  & 51.40$^\dagger$ & 51.98 & 52.05 \\
        \bottomrule
    \end{tabular}
    \caption{\textbf{Standard-gallery R@1 of the compared models}: published, with the official evaluation script (Official, for the released Sample4Geo checkpoints as stored with the weights, --: not run), and with our evaluation, which \cref{tab:dense_drop} uses. Trained models: last epoch. $^\dagger$Reported by AuxGeo~\cite{xia2025cross}.}
    \label{tab:supp_repro}
\end{table}

\section{Neighbor share of an ideal plateau}
\label{sec:supp_share}
We derive how often the best crop lies in a neighboring tile if the mean similarity is the plateau of \cref{eq:plateau} and the residuals favor none of the crops on it. We measure positions in units of the grid step $g$ and place the center of the positive tile at the origin, so the camera $\mathbf{x}$ is uniform over the cell $[-\tfrac12, \tfrac12]^2$, see \cref{sec:supp_cell}. Along each axis, the crop centers of all tiles form a lattice of stride $1/n$, so the plateau, the square of side 1 around $\mathbf{x}$, contains $n$ of them per axis and $n^2$ in total. The best crop is one of these $n^2$, chosen uniformly at random, so it lies in the positive tile with probability $\mathbb{E}[K]/n^2$, where $K$ is the number of the positive tile's crops on the plateau.

Along one axis, the crops of the positive tile lie at $\nu/n$ for the entries $\nu = -\tfrac{n-1}{2}, \dots, \tfrac{n-1}{2}$ of \cref{eq:lattice}. Crop $\nu$ lies on the plateau if $|\nu/n - x_1| \le \tfrac12$, which for $x_1$ uniform over $[-\tfrac12, \tfrac12]$ happens with probability $1 - |\nu|/n$. For odd $n$, $\sum_\nu |\nu|/n = (n^2 - 1)/(4n)$, so on average
\begin{equation}
    \sum_{\nu} \big(1 - |\nu|/n\big) = n \Big(\frac{3}{4} + \frac{1}{4n^2}\Big)
\end{equation}
of the tile's crops lie on the plateau along one axis. A crop lies on the plateau if it does so along both axes, and the two coordinates of the camera are independent, so $\mathbb{E}[K] = n^2 \big(\tfrac34 + \tfrac{1}{4n^2}\big)^2$, and the best crop lies in a neighboring tile with probability
\begin{equation}
    1 - \Big(\frac{3}{4} + \frac{1}{4n^2}\Big)^2.
    \label{eq:supp_share}
\end{equation}
This is zero for the standard gallery, $n = 1$, where the plateau contains only the positive tile, 39.5\% for $n = 3$ and 42.2\% for $n = 5$, and it approaches $7/16 = 43.8\%$ for fine lattices. In this limit, it is the average share of the plateau's area that lies outside the camera's cell: along each axis, the plateau and the cell overlap by $1 - |x_1|$, on average $3/4$. A Monte Carlo simulation with 400\,000 cameras gives the same values.

\section{Measuring the plateau}
\label{sec:supp_simabl}
This section details the measurements of \cref{sec:plateau}: the fit of the mean similarity, the 14 cases, the order of crop pairs, the residuals of nearby crops, a simulation, the ablation of the positive, and the threshold of the radius.

\paragraph{Fit of the mean similarity.} The level $a_q$ of each query and the mean similarity $\bar s$ of \cref{eq:score_model} are the least-squares fit
\begin{equation}
    (a, \bar s) = \operatorname*{arg\,min}_{a,\,\bar s} \sum_{q,r} \big(s(q, r) - a_q - \bar s(\boldsymbol{\delta}_r)\big)^2,
    \label{eq:fit}
\end{equation}
where $\bar s$ has one value per bin of the offset. At the minimum, $\bar s$ is the average of $s(q, r) - a_q$ over the pairs in each bin, $a_q$ is the average of $s(q, r) - \bar s(\boldsymbol{\delta}_r)$ over the references of query $q$, and the residual $e_{q,r}$ has zero mean in every bin and for every query. We compute the fit by alternating between the two averages until they converge. For the profiles of \cref{fig:score_field}b and for all numbers below, $\bar s$ depends on the distance $\rho = \|\boldsymbol{\delta}\|_2$ from the camera alone, in bins of $0.05g$ ($0.025g$ for LaSo), and the fit uses the crops with $\rho < 1.5g$, 72\% of the crops of a query. Fitting with all crops changes the drop of $\bar s$ at $0.3g$ and at $0.5g$ by at most 0.001. The standard deviation $\sigma$ is that of all residuals with $\rho \le g$, pooled over the queries. The maps of \cref{fig:score_field}a show $s(q, r) - a_q$, with $a_q$ from this fit, averaged per bin of $0.1g \times 0.1g$ of the offset.

\paragraph{Cases.} Each case uses the same 5\,000 test queries per split, drawn at random without replacement, and for each query the crops of the positive and its eight neighboring tiles, 225 crops or fewer at the border of the covered area. \Cref{tab:supp_cases} lists the 14 cases. Their best crop lies at a median distance of 0.26--0.37$g$ from the camera, against $0.40g$ on the ideal plateau. Over bootstrap samples of the queries, $\sigma$ has a standard error of at most 0.0004 and the share of neighbor misses one of at most 0.7 points.
\begin{table}[h]
    \centering
    \small
    \setlength{\tabcolsep}{3pt}
    \resizebox{\columnwidth}{!}{%
    \begin{tabular}{@{}llccccc@{}}
        \toprule
        & & & \multicolumn{2}{c}{Drop of $\bar s$ at} & Neighbor & \\
        \cmidrule(lr){4-5}
        Split & Model & $\sigma$ & $0.3g$ & $0.5g$ & misses [\%] & $\rho_{05}$ [$g$] \\
        \midrule
        VIGOR cross & Sample4Geo~\cite{deuser2023sample4geo} & 0.083 & 0.0169 & 0.0535 & 38.1 & 0.487 \\
                    & UnifyGeo~\cite{song2025unified}        & 0.069 & 0.0127 & 0.0427 & 38.5 & 0.531 \\
                    & Whole tile, public                     & 0.073 & 0.0183 & 0.0577 & 35.4 & 0.470 \\
                    & Whole tile, CC                         & 0.074 & 0.0213 & 0.0639 & 34.7 & 0.446 \\
        \midrule
        VIGOR same  & Sample4Geo~\cite{deuser2023sample4geo} & 0.089 & 0.0256 & 0.0762 & 32.7 & 0.414 \\
                    & UnifyGeo~\cite{song2025unified}        & 0.071 & 0.0193 & 0.0618 & 32.8 & 0.461 \\
                    & Whole tile, CC                         & 0.077 & 0.0274 & 0.0816 & 30.6 & 0.399 \\
        \midrule
        DReSS cross & Sample4Geo~\cite{deuser2023sample4geo} & 0.083 & 0.0195 & 0.0610 & 45.1 & 0.466 \\
                    & AuxGeo~\cite{xia2025cross}             & 0.082 & 0.0110 & 0.0513 & 46.0 & 0.497 \\
                    & Whole tile, public                     & 0.076 & 0.0165 & 0.0541 & 44.0 & 0.486 \\
                    & Whole tile, CC                         & 0.082 & 0.0150 & 0.0586 & 42.4 & 0.477 \\
        \midrule
        DReSS same  & Sample4Geo~\cite{deuser2023sample4geo} & 0.093 & 0.0219 & 0.0761 & 36.2 & 0.429 \\
                    & AuxGeo~\cite{xia2025cross}             & 0.089 & 0.0115 & 0.0615 & 39.6 & 0.478 \\
                    & Whole tile, CC                         & 0.088 & 0.0175 & 0.0715 & 36.4 & 0.445 \\
        \bottomrule
    \end{tabular}}
    \caption{\textbf{The 14 cases of \cref{sec:plateau}}: standard deviation $\sigma$ of the residual, drop of the mean similarity $\bar s$ from the camera to a distance of $0.3g$ and $0.5g$, share of queries whose best crop among the crops of the positive and its eight neighbors belongs to a neighboring tile, and plateau radius $\rho_{05}$, at which $\bar s$ has fallen by 0.05. Whole tile: our ViT-B with public initialization or with CC pre-training.}
    \label{tab:supp_cases}
\end{table}

\paragraph{Order of crop pairs.} Observation (i) of \cref{sec:plateau} rests on a direct count. For every query, we compare the crop nearest to the camera with every other crop and count how often the nearest crop has the higher similarity (\cref{tab:supp_pairs}). No ties occur. For the whole-tile models, the nearest crop wins against a crop $0.2$--$0.3g$ from the camera in 57--70\% of the pairs, and against a crop $0.4$--$0.5g$ away, near the border of $\mathcal{S}$, in 72--87\%. For LaSo, the shares are 97--98\% and 99\%.
\begin{table*}[t]
    \centering
    \footnotesize
    \setlength{\tabcolsep}{5pt}
    \begin{tabular}{@{}ll ccccccccc@{}}
        \toprule
        & & \multicolumn{9}{c}{Distance of the other crop from the camera [$g$]} \\
        \cmidrule(lr){3-11}
        Split & Model & 0.1--0.2 & 0.2--0.3 & 0.3--0.4 & 0.4--0.5 & 0.5--0.6 & 0.6--0.7 & 0.7--0.8 & 0.8--0.9 & 0.9--1.0 \\
        \midrule
        VIGOR cross & Sample4Geo~\cite{deuser2023sample4geo} & 56.4 & 65.2 & 71.6 & 80.1 & 87.0 & 93.1 & 96.0 & 97.5 & 98.1 \\
                    & UnifyGeo~\cite{song2025unified}        & 54.3 & 63.0 & 69.6 & 78.1 & 85.8 & 93.0 & 96.2 & 97.6 & 98.2 \\
                    & Whole tile, public                     & 55.2 & 63.2 & 71.0 & 80.3 & 87.2 & 93.0 & 96.3 & 98.0 & 98.5 \\
                    & Whole tile, CC                         & 55.8 & 64.7 & 73.3 & 82.1 & 88.9 & 94.6 & 97.4 & 98.7 & 99.0 \\
                    & LaSo, public                           & 82.0 & 97.4 & 98.5 & 99.1 & 99.2 & 99.5 & 99.4 & 99.6 & 99.6 \\
        \midrule
        VIGOR same  & Sample4Geo~\cite{deuser2023sample4geo} & 57.1 & 69.4 & 76.8 & 85.9 & 92.3 & 96.4 & 98.1 & 98.8 & 99.2 \\
                    & UnifyGeo~\cite{song2025unified}        & 57.0 & 67.9 & 75.3 & 84.0 & 92.0 & 96.6 & 98.3 & 98.9 & 99.1 \\
                    & Whole tile, CC                         & 57.6 & 69.5 & 77.8 & 87.1 & 93.2 & 97.1 & 98.6 & 99.2 & 99.4 \\
        \midrule
        DReSS cross & Sample4Geo~\cite{deuser2023sample4geo} & 54.7 & 61.5 & 67.6 & 76.3 & 84.5 & 90.6 & 93.3 & 94.8 & 95.4 \\
                    & AuxGeo~\cite{xia2025cross}             & 51.3 & 56.6 & 62.6 & 72.4 & 83.3 & 90.7 & 93.4 & 94.7 & 95.1 \\
                    & Whole tile, public                     & 53.5 & 60.9 & 66.7 & 75.6 & 84.1 & 90.8 & 93.5 & 95.1 & 95.7 \\
                    & Whole tile, CC                         & 53.1 & 59.5 & 65.4 & 75.0 & 85.5 & 92.6 & 95.4 & 96.5 & 97.0 \\
                    & LaSo, public                           & 83.3 & 98.0 & 98.5 & 98.9 & 99.0 & 99.1 & 99.2 & 99.2 & 99.3 \\
        \midrule
        DReSS same  & Sample4Geo~\cite{deuser2023sample4geo} & 55.5 & 65.5 & 72.3 & 82.6 & 90.5 & 95.2 & 96.7 & 97.6 & 98.0 \\
                    & AuxGeo~\cite{xia2025cross}             & 53.2 & 58.2 & 66.1 & 76.9 & 88.8 & 94.8 & 96.9 & 97.7 & 98.0 \\
                    & Whole tile, CC                         & 54.8 & 61.6 & 69.3 & 79.8 & 89.6 & 95.5 & 97.4 & 98.3 & 98.4 \\
        \bottomrule
    \end{tabular}
    \caption{\textbf{How often the crop nearest to the camera has a higher similarity than another crop}, in \% of the pairs, by the distance of the other crop from the camera. 50\% is chance. Whole tile and LaSo: our ViT-B with public initialization or with CC pre-training.}
    \label{tab:supp_pairs}
\end{table*}

\paragraph{Residuals of nearby crops.} The residuals of nearby crops are correlated, since the crops share most of their imagery. For two crops that are one stride apart and both lie within $0.5g$ of the camera, the correlation is 0.83--0.90, and their residuals differ with a standard deviation of only 0.38--0.48\,$\sigma$, against $1.41\,\sigma$ for independent residuals. At a separation of $0.45g$, it is 0.65--0.79\,$\sigma$. Measured against $\sigma$, the residual therefore appears larger than it is between nearby crops, which is why observation (i) counts pairs instead of comparing the drop of $\bar s$ with $\sigma$.

\begin{figure}[t]
    \centering
    \includegraphics[width=\linewidth]{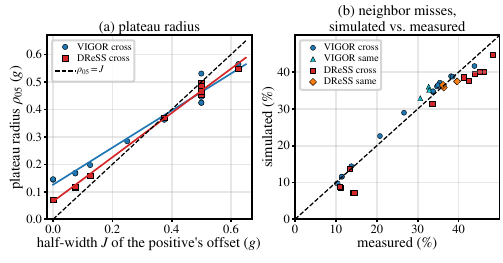}
    \caption{\textbf{Plateau radius and best crop of the whole-tile models and the ablation.} (a)~Plateau radius $\rho_{05}$ against the half-width $J$ of the positive's offset on the cross-area splits, with the linear fit per dataset. Dashed: $\rho_{05} = J$. (b)~Share of queries whose best crop lies in a neighboring tile, simulated against measured, for the same models and the whole-tile models on the same-area splits. Dashed: equality.}
    \label{fig:supp_plateau}
\end{figure}
\paragraph{Simulation.} The simulation tests whether the mean similarity and the residual we measure suffice to explain where the best crop lands. For each model, it takes three measured quantities: the mean similarity $\bar s(\rho)$ as a function of the distance $\rho$ from the camera, the standard deviation $\sigma$ of the residual, and the correlation of the residual between two crops at distance $h$, to which we fit $A\,e^{-h/\ell}$ for $h > 0$. It draws 4\,000 cameras uniformly in the cell, gives each of the 225 crops of the $3 \times 3$ tiles around the camera the similarity $\bar s(\rho) + \eta$, where $\eta$ is a Gaussian field with standard deviation $\sigma$ and the fitted correlation, and takes the best crop. Nothing is fitted to the share it predicts. As \cref{fig:supp_plateau}b shows, the simulated share follows the measured one over the whole-tile models and the ablation below, 34 models from 10 to 48\%, with a correlation of 0.98 and a mean deviation of 2.3 points. On DReSS, the simulation falls short by up to 6 points for the whole-tile models, and by 7 points for $J = 0$, whose peak is so narrow that the bins of $0.05g$ in which we measure $\bar s(\rho)$ resolve it too coarsely. The fit bounds $A$ by 1 and reaches this bound in all 14 cases, and the measured residual is neither stationary nor isotropic. The simulation is therefore a consistency check of the explanation, not a model of the residual.

\paragraph{Ablation of the positive.} For observation (iii), we change only the training positive. Instead of the official tile, the positive is a crop of the same size cut from the mosaic of the training tiles and centered at the camera plus an offset that is drawn uniformly from $[-J, J]^2$ for every training pair. We train $J = 0$, $0.075$, $0.125$, $0.25$, $0.375$, $0.5$ and $0.625\,g$ on VIGOR cross and $J = 0$, $0.075$, $0.125$, $0.375$ and $0.625\,g$ on DReSS cross, and, as a control, the official tile, whose offset from the camera is uniform over the cell, \ie, $J = g/2$. All arms fine-tune the same ViT-B for five epochs with the same schedule, starting from the model with CC pre-training, see \cref{sec:experiments}. The batches are built in the same way for every arm: each tile appears at most once per batch and no reference is held out, so the arms differ only in the positive. R@1 is the mean over three seeds, or of one seed where \cref{tab:supp_jmiss} marks it. For the plateau, the arms with $J = 0$ and $0.075\,g$ and the control are measured on a second seed as well. We measure the plateau's radius $\rho_{05}$ as the distance from the camera at which $\bar s(\rho)$ has fallen 0.05 below its value at the camera. As \cref{fig:supp_plateau}a shows, $\rho_{05}$ grows linearly with $J$ on both datasets: $\rho_{05} = 0.126\,g + 0.674\,J$ on VIGOR and $\rho_{05} = 0.065\,g + 0.806\,J$ on DReSS, with a root-mean-square deviation of $0.026\,g$ and $0.013\,g$. The fits include the whole-tile models of \cref{tab:supp_cases} on the cross-area splits at $J = g/2$. The slope stays below 1, its value for the ideal step of \cref{eq:plateau}, and the radius does not vanish at $J = 0$: a trained model rounds the edge of the plateau and adds a width of its own, 4.6\,m on VIGOR and 5 to 7\,m on DReSS. Although the model starts from a peak, five epochs suffice for the radius to follow $J$. The jittered positive with $J = g/2$ and the official tile, whose offsets from the camera follow the same distribution, give nearly the same radius on VIGOR, $0.46\,g$ and $0.42\,g$, so the plateau follows the range of the positive's offset, not whether the positive lies on the grid. The repeated arms reproduce the radius within $0.002\,g$.

\paragraph{Threshold of the radius.} The radius $\rho_{05}$ uses a drop of 0.05 in $\bar s$. \Cref{tab:supp_threshold} repeats the linear fit for other thresholds on the same models, including thresholds relative to the $\sigma$ of each model. The radius grows linearly with $J$ for every threshold, with a correlation above 0.97. Only the slope and the intercept depend on the threshold, so the values stated for $\rho_{05}$ hold for 0.05 only. The residual itself grows with $J$, on DReSS from $\sigma = 0.067$ at $J = 0$ to 0.094 at $J = 0.375g$, which steepens the slopes for the thresholds relative to $\sigma$.
\begin{table}[h]
    \centering
    \small
    \begin{tabular}{@{}lcc@{}}
        \toprule
        Threshold & VIGOR cross & DReSS cross \\
        \midrule
        0.02 & $0.08\,g + 0.46\,J$ & $0.04\,g + 0.57\,J$ \\
        0.05 & $0.13\,g + 0.67\,J$ & $0.06\,g + 0.81\,J$ \\
        0.10 & $0.19\,g + 0.86\,J$ & $0.10\,g + 0.98\,J$ \\
        \midrule
        $0.5\,\sigma$ & $0.10\,g + 0.60\,J$ & $0.05\,g + 0.78\,J$ \\
        $1.0\,\sigma$ & $0.15\,g + 0.79\,J$ & $0.08\,g + 0.95\,J$ \\
        \bottomrule
    \end{tabular}
    \caption{\textbf{Linear fit of the plateau radius against the half-width $J$ of the positive's offset} for different thresholds of the drop in $\bar s$, on the models of \cref{fig:supp_plateau}a.}
    \label{tab:supp_threshold}
\end{table}

\section{Readout rules on the dense index}
\label{sec:supp_rules}
\Cref{tab:supp_rules} compares the readout rules of \cref{sec:vote} on the dense index. Per-tile rules rate a tile by its best crop (max), the mean of its 25 crops (mean), the mean of its three best crops (top-3) or $\kappa \log \sum_r \exp(s(q, r)/\kappa)$ over its crops (log-sum-exp). The tile vote lets the 200 best crops of the query vote for their tile with the similarity weights $\exp((s(q, r) - s_{\max})/\kappa)$ of \cref{eq:vote}, and the spatial vote is \cref{eq:vote}. Together with their parameters, $\kappa \in \{0.003, 0.01, 0.03, 0.1, 0.3, 1\}$ for the log-sum-exp, $\kappa \in \{0.01, 0.03, 0.1, \infty\}$ for the tile vote, and $\kappa \in \{0.01, 0.03, 0.1, 0.3, 1\}$ and $b \in \{0.5, 1, 2, 4, 8\}$ strides for the spatial vote, these are the 38 settings among which \cref{sec:vote} selects. We select by the mean R@1 of all models of \cref{tab:dense_drop} on one benchmark and apply the choice to the other. Both directions select the spatial vote with $\kappa = 0.1$ and $b = 2$ strides. Averaging all crops of a tile helps on VIGOR, but on DReSS it changes R@1 by only $-2.7$ to $+0.4$ points. Log-sum-exp and the tile vote, which weight the crops by their similarity, gain more, and the spatial vote, which also pools crops across tile borders, is best for every model and split, each time with the selected setting.
\begin{table}[h]
    \centering
    \small
    \setlength{\tabcolsep}{2.6pt}
    \resizebox{\columnwidth}{!}{%
    \begin{tabular}{@{}ll cccccc@{}}
        \toprule
        & & & & & Log-sum- & Tile & Spatial \\
        Model & Split & Max & Mean & Top-3 & exp$^\ast$ & vote$^\ast$ & vote$^\ast$ \\
        \midrule
        Sample4Geo~\cite{deuser2023sample4geo} & VIGOR cross & 53.18 & 58.51 & 57.77 & 64.59 & 64.42 & \textbf{66.06} \\
                                               & VIGOR same  & 66.89 & 75.74 & 72.55 & 79.84 & 79.79 & \textbf{80.55} \\
                                               & DReSS cross & 28.65 & 27.50 & 32.24 & 34.61 & 34.89 & \textbf{36.39} \\
                                               & DReSS same  & 50.54 & 47.88 & 55.19 & 58.42 & 58.72 & \textbf{61.29} \\
        UnifyGeo~\cite{song2025unified}        & VIGOR cross & 51.69 & 57.23 & 56.75 & 64.13 & 64.03 & \textbf{66.12} \\
                                               & VIGOR same  & 65.70 & 75.39 & 71.55 & 79.77 & 79.64 & \textbf{80.94} \\
        AuxGeo~\cite{xia2025cross}             & DReSS cross & 29.11 & 28.74 & 33.19 & 36.52 & 37.31 & \textbf{39.38} \\
                                               & DReSS same  & 47.83 & 47.78 & 53.20 & 59.15 & 59.68 & \textbf{62.77} \\
        Whole tile (ours)                      & VIGOR cross & 57.42 & 64.94 & 62.29 & 69.35 & 69.20 & \textbf{70.75} \\
                                               & VIGOR same  & 67.88 & 76.58 & 73.10 & 79.88 & 79.84 & \textbf{80.65} \\
                                               & DReSS cross & 32.96 & 33.32 & 36.97 & 40.09 & 40.91 & \textbf{42.99} \\
                                               & DReSS same  & 49.48 & 49.17 & 54.36 & 58.67 & 59.25 & \textbf{61.83} \\
        \bottomrule
    \end{tabular}}
    \caption{\textbf{R@1 of the readout rules on the $5\times5$ dense index} for the models of \cref{tab:dense_drop}. $^\ast$Best setting of the rule's parameters on the test split. For the spatial vote, this is in every row the setting that \cref{sec:vote} selects across the benchmarks, $\kappa = 0.1$ and $b = 2$ strides. Whole tile: mean of three training runs. Best values in bold.}
    \label{tab:supp_rules}
\end{table}

\Cref{tab:supp_full} lists all three readouts for every model of \cref{tab:main}. The selected spatial vote lowers the R@1 of LaSo on every split, by 0.1 to 3.6 points, since the similarity of LaSo already peaks at the camera and averaging neighboring crops blurs this peak. Even a sharper kernel with $\kappa = 0.03$, selected on the test split, adds at most 0.7 points. With the best crop, the median localization error of LaSo ViT-L with CC pre-training is 3.25, 3.19, 8.52 and 7.62\,m on VIGOR cross, VIGOR same, DReSS cross and DReSS same, and the soft-argmax of \cref{sec:inference} lowers it to 1.52, 1.29, 3.46 and 2.61\,m. For the ViT-B with CC pre-training, the values are 3.36, 3.28, 9.06 and 7.90\,m and 1.72, 1.48, 4.19 and 3.13\,m. With 9 to 64 instead of 16 crops in the soft-argmax, these medians change by at most 0.07\,m.
\begin{table*}[t]
    \centering
    \footnotesize
    \setlength{\tabcolsep}{4.2pt}
    \begin{tabular}{@{}ll ccc ccc ccc ccc@{}}
        \toprule
        & & \multicolumn{3}{c}{VIGOR cross} & \multicolumn{3}{c}{VIGOR same} & \multicolumn{3}{c}{DReSS cross} & \multicolumn{3}{c}{DReSS same} \\
        \cmidrule(lr){3-5}\cmidrule(lr){6-8}\cmidrule(lr){9-11}\cmidrule(lr){12-14}
        Training & Backbone & Tiles & Dense & +Vote & Tiles & Dense & +Vote & Tiles & Dense & +Vote & Tiles & Dense & +Vote \\
        \midrule
        \multicolumn{14}{@{}l}{\emph{Public initialization, benchmark data only}} \\
        Whole tile & ViT-B & 64.32 & 57.42 & 70.75 & 76.96 & 67.88 & 80.65 & 33.37 & 32.96 & 42.99 & 51.03 & 49.48 & 61.83 \\
        LaSo       & ViT-B & 27.40 & 85.98 & 84.59 & 28.79 & 89.95 & 89.57 & 18.20 & 76.31 & 72.67 & 21.45 & 87.49 & 85.04 \\
        Whole tile & ViT-L & \textbf{72.99} & 64.67 & 77.64 & \textbf{81.35} & 71.69 & 83.88 & \textbf{46.93} & 46.09 & 56.46 & \textbf{64.00} & 60.32 & 72.40 \\
        LaSo       & ViT-L & 32.09 & \textbf{89.99} & \textbf{89.32} & 32.94 & \textbf{92.21} & \textbf{92.02} & 23.27 & \textbf{85.17} & \textbf{82.75} & 26.67 & \textbf{92.39} & \textbf{90.94} \\
        \midrule
        \multicolumn{14}{@{}l}{\emph{CC pre-training}} \\
        Whole tile & ViT-B & 71.67 & 61.42 & 76.60 & 78.64 & 67.83 & 81.91 & 43.21 & 40.42 & 54.47 & 55.57 & 51.90 & 66.66 \\
        LaSo       & ViT-B & 30.89 & 88.09 & 86.76 & 29.63 & 90.01 & 89.66 & 19.94 & 80.18 & 76.56 & 21.88 & 87.80 & 85.27 \\
        Whole tile & ViT-L & \textbf{77.98} & 64.88 & 81.26 & \textbf{82.65} & 69.51 & 84.53 & \textbf{55.70} & 50.68 & 65.55 & \textbf{66.97} & 59.16 & 74.86 \\
        LaSo       & ViT-L & 35.28 & \textbf{90.66} & \textbf{89.97} & 33.74 & \textbf{92.03} & \textbf{91.89} & 25.01 & \textbf{86.85} & \textbf{84.45} & 26.72 & \textbf{92.30} & \textbf{90.61} \\
        \bottomrule
    \end{tabular}
    \caption{\textbf{All readouts of the models of \cref{tab:main}}: R@1 on the standard gallery (Tiles), on the $5 \times 5$ dense index with the best crop (Dense) and with the spatial vote of \cref{sec:vote} (+Vote). Mean over three seeds for the ViT-B models and for LaSo ViT-L with CC pre-training, one seed otherwise. Best values per block in bold.}
    \label{tab:supp_full}
\end{table*}

\Cref{tab:supp_n3} evaluates every model of \cref{tab:main} also on a $3 \times 3$ index, which stores 9 instead of 25 embeddings per tile, with the best crop. LaSo, trained for $n = 5$, loses more R@1 on DReSS than on VIGOR, since the stride on DReSS grows to 24--36\,m. The whole-tile models reach a slightly higher R@1 on the $3 \times 3$ than on the $5 \times 5$ index, as the expected share of misses on their plateau is slightly smaller, 39.5 instead of 42.2\%, see \cref{sec:supp_share}, but stay below their standard-gallery R@1.
\begin{table*}[t]
    \centering
    \footnotesize
    \setlength{\tabcolsep}{4.2pt}
    \begin{tabular}{@{}ll ccc ccc ccc ccc@{}}
        \toprule
        & & \multicolumn{3}{c}{VIGOR cross} & \multicolumn{3}{c}{VIGOR same} & \multicolumn{3}{c}{DReSS cross} & \multicolumn{3}{c}{DReSS same} \\
        \cmidrule(lr){3-5}\cmidrule(lr){6-8}\cmidrule(lr){9-11}\cmidrule(lr){12-14}
        Training & Backbone & Tiles & $3 \times 3$ & $5 \times 5$ & Tiles & $3 \times 3$ & $5 \times 5$ & Tiles & $3 \times 3$ & $5 \times 5$ & Tiles & $3 \times 3$ & $5 \times 5$ \\
        \midrule
        \multicolumn{14}{@{}l}{\emph{Public initialization, benchmark data only}} \\
        Whole tile & ViT-B & 64.32 & 57.92 & 57.42 & 76.96 & 69.03 & 67.88 & 33.37 & 33.23 & 32.96 & 51.03 & 50.20 & 49.48 \\
        LaSo & ViT-B & 27.40 & 82.73 & 85.98 & 28.79 & 87.90 & 89.95 & 18.20 & 68.10 & 76.31 & 21.45 & 80.48 & 87.49 \\
        Whole tile & ViT-L & \textbf{72.99} & 65.74 & 64.67 & \textbf{81.35} & 72.95 & 71.69 & \textbf{46.93} & 46.63 & 46.09 & \textbf{64.00} & 61.44 & 60.32 \\
        LaSo & ViT-L & 32.09 & \textbf{88.65} & \textbf{89.99} & 32.94 & \textbf{91.15} & \textbf{92.21} & 23.27 & \textbf{80.03} & \textbf{85.17} & 26.67 & \textbf{89.01} & \textbf{92.39} \\
        \midrule
        \multicolumn{14}{@{}l}{\emph{CC pre-training}} \\
        Whole tile & ViT-B & 71.67 & 62.16 & 61.42 & 78.64 & 69.08 & 67.83 & 43.21 & 40.74 & 40.42 & 55.57 & 52.54 & 51.90 \\
        LaSo & ViT-B & 30.89 & 85.87 & 88.09 & 29.63 & 88.26 & 90.01 & 19.94 & 73.02 & 80.18 & 21.88 & 81.00 & 87.80 \\
        Whole tile & ViT-L & \textbf{77.98} & 66.17 & 64.88 & \textbf{82.65} & 70.80 & 69.51 & \textbf{55.70} & 51.52 & 50.68 & \textbf{66.97} & 60.46 & 59.16 \\
        LaSo & ViT-L & 35.28 & \textbf{89.67} & \textbf{90.66} & 33.74 & \textbf{91.28} & \textbf{92.03} & 25.01 & \textbf{82.47} & \textbf{86.85} & 26.72 & \textbf{88.68} & \textbf{92.30} \\
        \bottomrule
    \end{tabular}
    \caption{\textbf{R@1 on indexes of different density}: the standard gallery (Tiles) and dense indexes with $3 \times 3$ and $5 \times 5$ crops per tile, read out by the best crop, for the models of \cref{tab:main}. LaSo is trained for $n = 5$. Mean over three seeds for the ViT-B models and for LaSo ViT-L with CC pre-training, one seed otherwise. Best values per block in bold.}
    \label{tab:supp_n3}
\end{table*}

\section{Types of misses on the dense index}
\label{sec:supp_errors}
\Cref{tab:supp_r5} lists R@5 and the hit rate of the whole-tile models of \cref{tab:dense_drop} on the standard gallery and the dense index, and \cref{tab:supp_errtypes} splits the misses by how far the retrieved tile lies from the positive. On the dense index, R@5 rises for every model, and the hit rate for every model except AuxGeo, where it falls by 0.8 and 3.0 points. The misses move to the neighboring tiles. Their share rises by 10.0--17.7 points for every whole-tile model on every split, on VIGOR from 11.7--16.3\% to 28.7--30.4\% and on DReSS from 7.3--11.7\% to 17.8--28.0\%, while misses four or more grid steps away fall by 2.0--11.0 points. On DReSS, this fall offsets most of the new neighbor misses, which is why the R@1 loss is smaller there. LaSo keeps the neighbor share at 4.7--7.3\% on the dense index.

\begin{table}[h]
    \centering
    \small
    \setlength{\tabcolsep}{3pt}
    \resizebox{\columnwidth}{!}{%
    \begin{tabular}{@{}llcccc@{}}
        \toprule
        & & \multicolumn{2}{c}{R@5} & \multicolumn{2}{c}{Hit rate} \\
        \cmidrule(lr){3-4}\cmidrule(lr){5-6}
        Model & Split & Tiles & Dense & Tiles & Dense \\
        \midrule
        Sample4Geo~\cite{deuser2023sample4geo} & VIGOR cross & 83.4 & 87.0 & 74.5 & 81.1 \\
                                               & VIGOR same  & 95.7 & 97.8 & 93.7 & 96.2 \\
                                               & DReSS cross & 49.8 & 56.2 & 31.7 & 33.1 \\
                                               & DReSS same  & 73.4 & 83.0 & 56.4 & 58.0 \\
        UnifyGeo~\cite{song2025unified}        & VIGOR cross & 81.6 & 86.0 & 72.5 & 79.9 \\
                                               & VIGOR same  & 94.9 & 97.3 & 92.5 & 95.6 \\
        AuxGeo~\cite{xia2025cross}             & DReSS cross & 53.3 & 58.5 & 34.6 & 33.8 \\
                                               & DReSS same  & 75.9 & 83.1 & 58.4 & 55.4 \\
        \midrule
        Whole tile (ours)                      & VIGOR cross & 85.6 & 91.2 & 78.4 & 86.3 \\
                                               & VIGOR same  & 94.7 & 97.7 & 92.2 & 96.0 \\
                                               & DReSS cross & 53.7 & 62.2 & 36.2 & 38.2 \\
                                               & DReSS same  & 71.9 & 82.4 & 55.3 & 57.1 \\
        \bottomrule
    \end{tabular}}
    \caption{\textbf{R@5 and hit rate} (top-1 tile is the positive or a semi-positive) of the models in \cref{tab:dense_drop}, on the standard gallery (Tiles) and the $5\times5$ dense index with the best crop (Dense). Semi-positives as labeled by the benchmarks; on DReSS, 74\% of the queries have none. The official code of Sample4Geo uses a stricter definition, which counts a query only if no tile other than a semi-positive outranks the positive; with it, Sample4Geo reaches 69.86 / 89.94 on the Tiles of VIGOR cross / same (published 69.87 / 89.82). Whole tile: mean of three training runs.}
    \label{tab:supp_r5}
\end{table}

\begin{table*}[t]
    \centering
    \footnotesize
    \setlength{\tabcolsep}{5pt}
    \begin{tabular}{@{}lll cccc cccc@{}}
        \toprule
        & & & \multicolumn{4}{c}{Cross-area} & \multicolumn{4}{c}{Same-area} \\
        \cmidrule(lr){4-7}\cmidrule(lr){8-11}
        Benchmark & Model & Index & Neighbor & Near & Far & Other city & Neighbor & Near & Far & Other city \\
        \midrule
        VIGOR & Sample4Geo~\cite{deuser2023sample4geo} & Tiles & 13.8 & 1.6 & 20.6 & 2.6 & 16.3 & 1.0 & 4.1 & 0.7 \\
              &                                        & Dense & 28.7 & 1.2 & 15.0 & 1.9 & 29.9 & 0.7 & 2.1 & 0.4 \\
              & UnifyGeo~\cite{song2025unified}        & Tiles & 11.7 & 1.3 & 22.1 & 3.5 & 14.5 & 1.0 & 5.0 & 1.0 \\
              &                                        & Dense & 28.8 & 1.1 & 15.9 & 2.5 & 30.4 & 0.7 & 2.6 & 0.5 \\
              & Whole tile (ours)                      & Tiles & 14.8 & 1.3 & 17.3 & 2.3 & 15.8 & 0.9 & 5.3 & 1.0 \\
              &                                        & Dense & 30.2 & 0.9 & 10.5 & 1.3 & 28.7 & 0.6 & 2.2 & 0.5 \\
              & LaSo                                   & Tiles & 2.6 & 2.4 & 55.6 & 11.9 & 3.1 & 2.7 & 50.7 & 14.8 \\
              &                                        & Dense & 7.2 & 0.6 & 5.3 & 0.9 & 7.3 & 0.5 & 1.8 & 0.5 \\
        \midrule
        DReSS & Sample4Geo~\cite{deuser2023sample4geo} & Tiles & 7.8 & 3.2 & 45.9 & 13.7 & 11.7 & 3.2 & 29.6 & 3.5 \\
              &                                        & Dense & 17.8 & 2.9 & 39.6 & 11.0 & 25.4 & 2.4 & 19.9 & 1.7 \\
              & AuxGeo~\cite{xia2025cross}             & Tiles & 7.3 & 3.4 & 45.4 & 11.6 & 10.3 & 3.3 & 28.8 & 3.3 \\
              &                                        & Dense & 19.5 & 2.9 & 39.1 & 9.4 & 28.0 & 2.4 & 20.0 & 1.8 \\
              & Whole tile (ours)                      & Tiles & 8.4 & 2.9 & 47.5 & 7.8 & 10.8 & 2.7 & 32.1 & 3.4 \\
              &                                        & Dense & 20.2 & 2.5 & 38.4 & 6.0 & 25.9 & 1.9 & 21.1 & 1.5 \\
              & LaSo                                   & Tiles & 2.7 & 2.5 & 58.3 & 18.2 & 3.1 & 2.7 & 56.2 & 16.6 \\
              &                                        & Dense & 5.3 & 0.8 & 13.7 & 3.9 & 4.7 & 0.5 & 6.5 & 0.7 \\
        \bottomrule
    \end{tabular}
    \caption{\textbf{Misses by type}, in \% of all queries, on the standard gallery (Tiles) and the $5\times5$ dense index with the best crop (Dense): the retrieved tile is one of the eight tiles around the positive (neighbor), two or three grid steps away (near), four or more steps away in the same city (far), or in another test city. Whole tile and LaSo: ViT-B with public initialization, one training run on VIGOR and the mean of three on DReSS.}
    \label{tab:supp_errtypes}
\end{table*}

\section{Decentrality on DReSS}
\label{sec:supp_decentrality}
\Cref{tab:decentrality} reports R@1 on the four subsets S1 to S4 into which DReSS divides its test queries by the distance of the camera from the center of its positive~\cite{xia2025cross}. On the standard gallery, the R@1 of the whole-tile models falls by about half from S1 to S4. The spatial vote on the dense index reduces this fall but does not remove it. LaSo stays within one point from S1 to S3 on the dense index and loses 9--10 points in S4, where crops of neighboring tiles can be almost as close to the camera as the positive. On the standard gallery, LaSo reaches 89.0 and 94.9 in S1, 3--5 points more than on the dense index, since the tile center then lies on its peak, and falls to 2.1 in S4.
\begin{table}[h]
    \centering
    \footnotesize
    \setlength{\tabcolsep}{2.2pt}
    \resizebox{\columnwidth}{!}{%
    \begin{tabular}{@{}ll cccc cccc@{}}
        \toprule
        & & \multicolumn{4}{c}{DReSS cross} & \multicolumn{4}{c}{DReSS same} \\
        \cmidrule(lr){3-6}\cmidrule(lr){7-10}
        Method & Index & S1 & S2 & S3 & S4 & S1 & S2 & S3 & S4 \\
        \midrule
        Sample4Geo & Tiles & 43.0 & 40.6 & 32.4 & 19.4 & 73.6 & 69.1 & 58.1 & 36.1 \\
        Sample4Geo & Dense +Vote & 44.7 & 44.1 & 39.4 & 29.2 & 74.4 & 73.3 & 67.2 & 49.1 \\
        AuxGeo & Tiles & 44.4 & 42.7 & 36.5 & 22.2 & 72.4 & 69.5 & 61.1 & 39.3 \\
        AuxGeo & Dense +Vote & 48.0 & 46.7 & 43.0 & 31.8 & 75.4 & 74.1 & 68.4 & 51.1 \\
        Whole tile & Tiles & 59.6 & 56.3 & 48.3 & 30.6 & 75.6 & 71.5 & 62.6 & 39.5 \\
        Whole tile & Dense +Vote & 66.1 & 65.1 & 59.8 & 43.6 & 80.2 & 78.9 & 73.2 & 53.8 \\
        LaSo & Tiles & \textbf{89.0} & 54.2 & 7.9 & 2.1 & \textbf{94.9} & 61.1 & 8.8 & 2.1 \\
        LaSo & Dense & 84.3 & \textbf{84.8} & \textbf{84.0} & \textbf{74.4} & 91.6 & \textbf{92.1} & \textbf{91.5} & \textbf{82.4} \\
        \bottomrule
    \end{tabular}}
    \caption{\textbf{R@1 by decentrality on DReSS}, on the subsets S1 to S4 of the test queries, whose cameras lie increasingly far from the center of their positive~\cite{xia2025cross}. Index as in \cref{tab:main}. Whole tile and LaSo: ViT-B with CC pre-training, mean over three seeds. Best values in bold.}
    \label{tab:decentrality}
\end{table}

\section{Cost of the dense index}
\label{sec:supp_cost}
The dense index with $n = 5$ stores 25 embeddings per tile, so its memory, its embedding time and the work of an exact search are 25 times those of the standard gallery. The smallest index, VIGOR cross-area with 1.16\,M crops, takes 1.8\,GB for ViT-B and 2.4\,GB for ViT-L in half precision, and the largest, DReSS same-area with 10.57\,M crops, 16.2 and 21.6\,GB. The crops are embedded once per gallery, which takes 1.2 and 3.3 V100 GPU-hours on VIGOR cross-area and 10.9 and 30.7 on DReSS same-area, including cutting the crops from the mosaic. An exact search compares the query with every crop and takes 0.12--0.13\,ms per query on VIGOR cross-area and 1.22\,ms on DReSS same-area, on one V100 GPU in half precision. The query itself is embedded once, as for the standard gallery. On VIGOR cross, two standard compressions keep R@1. Scalar quantization of the embeddings to int8 halves the memory and changes R@1 by at most 0.03 points. An inverted-file index with 2048 lists, of which the 32 nearest are searched~\cite{johnson2019billion}, compares a query with about 18\,k crops and 2\,k list centers, fewer than the 46.6\,k tiles of the standard gallery, and loses 0.5 points for LaSo ViT-B with public initialization and 0.4 points for LaSo ViT-L with pre-training, at 0.40 and 0.55\,ms per query on a CPU.

\section{Ablation of the positive with standard deviations}
\label{sec:supp_ablation}
\Cref{tab:supp_ablation} repeats \cref{tab:ablation} with the standard deviation over the three seeds.
\begin{table}[h]
    \centering
    \small
    \setlength{\tabcolsep}{2pt}
    \resizebox{\columnwidth}{!}{%
    \begin{tabular}{@{}l ccc@{}}
        \toprule
        Positive & $n{=}1$ & $n{=}3$ & $n{=}5$ \\
        \midrule
        \multicolumn{4}{@{}l}{\emph{VIGOR cross}} \\
        Whole tile & 69.03 $\pm$ 0.36 & 63.67 $\pm$ 0.38 & 63.04 $\pm$ 0.35 \\
        Camera, $J = 0$ & 30.61 $\pm$ 0.03 & 83.46 $\pm$ 0.03 & 86.60 $\pm$ 0.06 \\
        Camera, $J = 0.075g$ & 36.85 $\pm$ 0.48 & 84.74 $\pm$ 0.10 & 85.76 $\pm$ 0.16 \\
        Camera, $J = 0.125g$ & 45.20 $\pm$ 0.63 & 84.41 $\pm$ 0.14 & 83.82 $\pm$ 0.10 \\
        Camera, $0.075g$, 2 adjacent, soft & 29.99 $\pm$ 0.14 & 84.52 $\pm$ 0.16 & 87.59 $\pm$ 0.07 \\
        Lattice, no siblings & 39.78 $\pm$ 0.12 & 84.66 $\pm$ 0.11 & 84.89 $\pm$ 0.04 \\
        Lattice, 2 of other tiles & 36.52 $\pm$ 0.06 & 85.22 $\pm$ 0.14 & 86.50 $\pm$ 0.11 \\
        Lattice, 2 random, hard & 24.42 $\pm$ 0.13 & 81.87 $\pm$ 0.10 & 87.09 $\pm$ 0.04 \\
        Lattice, 2 random, soft & 30.45 $\pm$ 0.23 & 84.81 $\pm$ 0.10 & 87.33 $\pm$ 0.07 \\
        LaSo: 2 adjacent, soft & 31.93 $\pm$ 0.17 & 84.96 $\pm$ 0.05 & 87.44 $\pm$ 0.05 \\
        \midrule
        \multicolumn{4}{@{}l}{\emph{DReSS cross}} \\
        Whole tile & 38.58 $\pm$ 0.21 & 39.10 $\pm$ 0.29 & 39.30 $\pm$ 0.29 \\
        Camera, $J = 0$ & 8.02 $\pm$ 0.08 & 36.87 $\pm$ 0.18 & 61.94 $\pm$ 0.20 \\
        Camera, $J = 0.075g$ & 15.40 $\pm$ 0.08 & 62.25 $\pm$ 0.04 & 75.88 $\pm$ 0.13 \\
        Camera, $J = 0.125g$ & 22.27 $\pm$ 0.05 & 69.12 $\pm$ 0.06 & 73.70 $\pm$ 0.10 \\
        Camera, $0.075g$, 2 adjacent, soft & 17.98 $\pm$ 0.07 & 67.68 $\pm$ 0.02 & 77.70 $\pm$ 0.01 \\
        Lattice, no siblings & 18.48 $\pm$ 0.14 & 66.92 $\pm$ 0.09 & 75.38 $\pm$ 0.02 \\
        Lattice, 2 of other tiles & 17.87 $\pm$ 0.09 & 66.64 $\pm$ 0.13 & 76.61 $\pm$ 0.04 \\
        Lattice, 2 random, hard & 13.22 $\pm$ 0.08 & 59.86 $\pm$ 0.24 & 76.88 $\pm$ 0.03 \\
        Lattice, 2 random, soft & 17.66 $\pm$ 0.10 & 67.87 $\pm$ 0.13 & 77.46 $\pm$ 0.18 \\
        LaSo: 2 adjacent, soft & 18.74 $\pm$ 0.03 & 68.42 $\pm$ 0.10 & 77.25 $\pm$ 0.08 \\
        \bottomrule
    \end{tabular}}
    \caption{\textbf{R@1 of the training positives of \cref{tab:ablation}}, mean $\pm$ standard deviation over three seeds. The seeds share the initialization and part of the batch order, so the standard deviations understate the full seed variance.}
    \label{tab:supp_ablation}
\end{table}

\begin{figure*}[t]
    \centering
    \includegraphics[width=\linewidth]{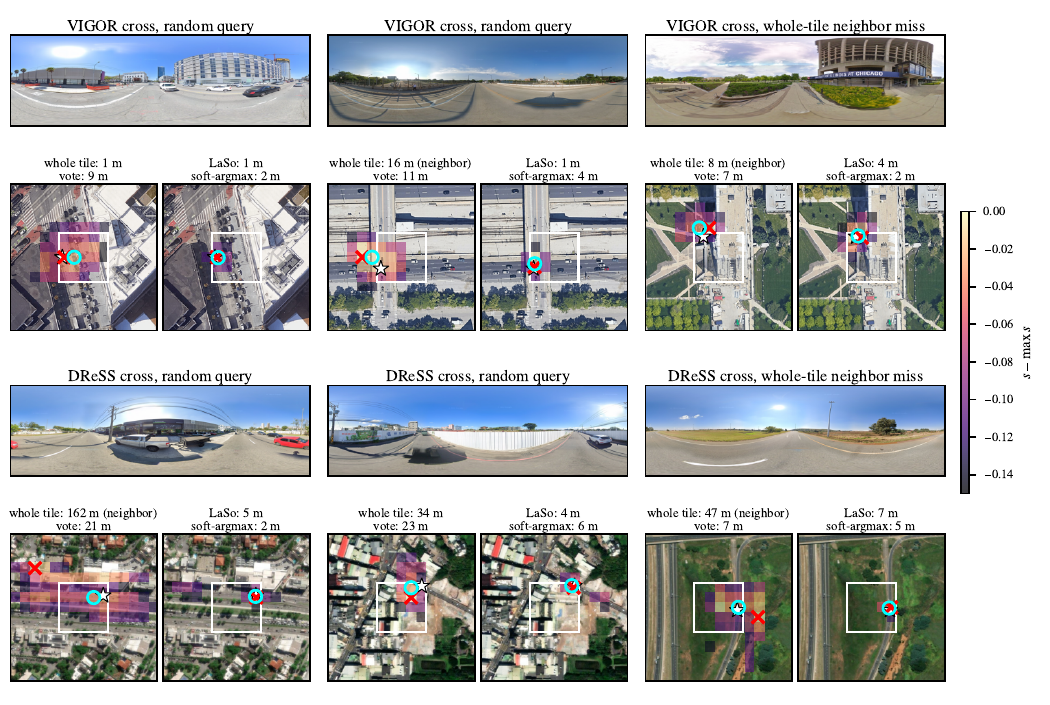}
    \caption{\textbf{Similarity maps of the whole-tile model and LaSo}, both ViT-B with CC pre-training, for random queries and for a query whose best whole-tile crop lies in a neighboring tile. Each map shows the $3 \times 3$ tiles around the positive with the similarity $s - \max s$ of their $5 \times 5$ crops, in cosine similarity and only for crops within 0.15 of the best. Star: camera. White square: cell of the positive. Red cross: best crop. Cyan circle: spatial vote of \cref{sec:vote} for the whole-tile model, soft-argmax of \cref{sec:inference} for LaSo. Titles: distance of the best crop and of the vote or soft-argmax from the camera. Neighbor, neighbor miss: the best crop lies in a neighboring tile.}
    \label{fig:supp_qualitative}
\end{figure*}

\section{Published results on the standard gallery}
\label{sec:supp_published}
\Cref{tab:supp_published} lists the published R@1 on the standard gallery next to LaSo on the dense index. The two use different indexes, so the table compares how well the methods locate the camera through retrieval, not the same protocol: the published models that we could evaluate lose R@1 on the dense index, see \cref{tab:dense_drop}, and LaSo reaches only 18 to 35\% on the standard gallery, see \cref{tab:supp_full}. The strongest published results on VIGOR use a 608\,M encoder with extra training data (GeoBridge++) or re-rank the 100 best tiles over 784 translations each (C-BEV). DReSS was introduced together with AuxGeo, and we know of no published retrieval results on it beyond those of the AuxGeo paper, which include Sample4Geo.
\begin{table}[h]
    \centering
    \small
    \setlength{\tabcolsep}{3pt}
    \resizebox{\columnwidth}{!}{%
    \begin{tabular}{@{}lllcccc@{}}
        \toprule
        Method & Params & Test stage & V-cross & V-same & D-cross & D-same \\
        \midrule
        \multicolumn{7}{@{}l}{\emph{Published, standard gallery}} \\
        Sample4Geo~\cite{deuser2023sample4geo} & 88.6\,M & 1 vector & 61.70 & 77.86 & 28.23$^\ast$ & 51.40$^\ast$ \\
        AuxGeo~\cite{xia2025cross} & 88.6\,M & 1 vector & 63.94 & 80.34 & \textbf{32.44} & \textbf{54.70} \\
        CV-Cities~\cite{huang2024cv} & 90.5\,M & 1 vector & 64.61 & 78.27 & -- & -- \\
        Geo$^2$~\cite{zhang2026geo} & 1.07\,B & 1 vector & 66.71 & 81.59 & -- & -- \\
        UnifyGeo~\cite{song2025unified} & 57.7\,M & re-rank top 5 & 67.58 & 82.80 & -- & -- \\
        EP-BEV~\cite{ye2024cross} & 177\,M & 2 models & 72.19 & 82.18 & -- & -- \\
        Warp-free~\cite{song2026warp} & 88\,M & 1 vector & 73.60 & 83.15 & -- & -- \\
        GeoBridge$^\dagger$~\cite{geobridge2026} & 608\,M & 1 vector & 73.87 & 85.82 & -- & -- \\
        GeoBridge++$^\dagger$~\cite{geobridgepp2025} & 608\,M & 1 vector & 78.20 & \textbf{89.49} & -- & -- \\
        C-BEV~\cite{fervers2023c} & -- & re-rank top 100 & \textbf{80.01} & 87.49 & -- & -- \\
        \midrule
        \multicolumn{7}{@{}l}{\emph{LaSo, dense index with $n = 5$}} \\
        ConvNeXt-B & 88.6\,M & 1 vector, dense & 81.64 & -- & 69.43 & -- \\
        ViT-B & 85.6\,M & 1 vector, dense & 85.98 & 89.95 & 76.31 & 87.49 \\
        ViT-B, CC pre-training$^\dagger$ & 85.6\,M & 1 vector, dense & 88.09 & 90.01 & 80.18 & 87.80 \\
        ViT-L & 303\,M & 1 vector, dense & 89.99 & \textbf{92.21} & 85.17 & \textbf{92.39} \\
        ViT-L, CC pre-training$^\dagger$ & 303\,M & 1 vector, dense & \textbf{90.66} & 92.03 & \textbf{86.85} & 92.30 \\
        \bottomrule
    \end{tabular}}
    \caption{\textbf{Published R@1 on the standard gallery and LaSo on the dense index.} Params: parameters at test time, derived from the stated backbones where the paper does not give them. --: not reported or not trained. $^\ast$As reported by AuxGeo. $^\dagger$Extra training data: multi-view data for GeoBridge, the CC pre-training of \cref{sec:setup} for LaSo. LaSo: as in \cref{tab:main}, mean over three seeds for ViT-B and for ViT-L with pre-training, else one seed. ConvNeXt-B: three seeds on VIGOR cross, one on DReSS cross. Dense: 25 embeddings per tile. V: VIGOR, D: DReSS. Best values per block in bold.}
    \label{tab:supp_published}
\end{table}

\section{Qualitative results}
\label{sec:supp_qualitative}
\Cref{fig:supp_qualitative} shows, for VIGOR cross and DReSS cross, two random test queries and one query drawn at random among those for which the best crop of the whole-tile model lies in a neighboring tile. The whole-tile model rates a broad region around the camera alike, as expected from the plateau of \cref{sec:plateau}, and its best crop can lie anywhere in this region. LaSo singles out one or two crops at the camera. For single queries, the soft-argmax can lie farther from the camera than the best crop.

\section{Training and index details}
\label{sec:supp_training}
\paragraph{Training.} We use one warm-up epoch on VIGOR and two on DReSS. Batches are built from geographic and visual neighbors as in Sample4Geo~\cite{deuser2023sample4geo}, re-mined every four epochs, or every six epochs during CC pre-training. The five-epoch fine-tunes of the ablations mine visual neighbors once. We augment with horizontal flips and rotations by multiples of $90^\circ$ and apply the corresponding transformation to the paired panorama. \Cref{sec:supp_lattice} compares LaSo trained with $n = 3$, 5 and 7. The ConvNeXt-B~\cite{liu2022convnet} models of \cref{tab:supp_published} are trained with a learning rate of $10^{-3}$, three seeds on VIGOR cross-area and one on DReSS cross-area.

\paragraph{Mosaic.} Every crop is cut from a mosaic of the gallery tiles at their georeferenced positions. The part of a crop that overlaps its tile keeps the pixels of that tile, and the part beyond its border comes from the neighboring tiles. Where the mosaic does not cover a crop, the missing region stays black, which affects about 1\% of the crops on VIGOR and 2\% on DReSS. The dense index has 1.16\,M crops on VIGOR cross-area and 10.6\,M on DReSS same-area (\cref{sec:supp_cost}).

\section{Range of the positive: plateau, R@1 and misses}
\label{sec:supp_jarms}
\Cref{tab:supp_jmiss} lists, for every arm of the ablations of \cref{sec:supp_simabl} and \cref{tab:ablation}, the plateau radius $\rho_{05}$, R@1 on the standard gallery and on the $5\times5$ index, and the misses on that index by type. On VIGOR, R@1 on the $5\times5$ index falls as the radius grows, from 86.6 at $J = 0$ to 52.2 at $J = 0.625g$, with a correlation of $-0.998$ over the whole tile and the seven camera-centered arms. On DReSS, the correlation is $-0.93$, since the arm with $J = 0$ has the narrowest peak but not the highest R@1 (\cref{sec:lattice_pos}). On the standard gallery, R@1 is highest near the whole-tile case, $J = 0.375g$ to $g/2$.

Neighbor misses grow with $J$, as the plateau does. The misses four or more grid steps away also depend on the positive: on DReSS cross, their share is 31.2\% for the whole tile, 13.0--13.4\% for the lattice positive and for $J = 0.075g$ and $0.125g$, and 25.0--39.4\% for larger $J$. At $J = 0$ it is 24.2\%, the case of \cref{sec:lattice_pos}. Siblings change it by at most 0.9 points.
\begin{table}[h]
    \centering
    \small
    \setlength{\tabcolsep}{2.5pt}
    \resizebox{\columnwidth}{!}{%
    \begin{tabular}{@{}l c cc cccc@{}}
        \toprule
        & & \multicolumn{2}{c}{R@1} & \multicolumn{4}{c}{Misses at $n = 5$ [\%]} \\
        \cmidrule(lr){3-4}\cmidrule(lr){5-8}
        Positive & $\rho_{05}$ [$g$] & $n{=}1$ & $n{=}5$ & Neighbor & Near & Far & Other city \\
        \midrule
        \multicolumn{8}{@{}l}{\emph{VIGOR cross}} \\
        Whole tile & 0.424 & 69.03 & 63.04 & 28.6 & 0.8 & 6.7 & 0.8 \\
        Whole tile, 2 of other tiles & 0.417 & 69.33 & 63.35 & 28.5 & 0.8 & 6.6 & 0.7 \\
        Camera, $J = 0$ & 0.145 & 30.61 & 86.60 & 8.0 & 0.5 & 4.2 & 0.8 \\
        Camera, $J = 0.075g$ & 0.168 & 36.85 & 85.76 & 9.1 & 0.5 & 3.9 & 0.7 \\
        Camera, $J = 0.125g$ & 0.197 & 45.20 & 83.82 & 11.1 & 0.5 & 3.9 & 0.7 \\
        Camera, $J = 0.25g$$^\ast$ & 0.284 & 64.85 & 76.66 & 17.7 & 0.6 & 4.5 & 0.6 \\
        Camera, $J = 0.375g$$^\ast$ & 0.362 & 70.59 & 70.11 & 23.1 & 0.7 & 5.4 & 0.7 \\
        Camera, $J = 0.5g$$^\ast$ & 0.460 & 67.83 & 61.43 & 29.6 & 0.9 & 7.3 & 0.8 \\
        Camera, $J = 0.625g$$^\ast$ & 0.566 & 61.39 & 52.15 & 36.6 & 1.1 & 9.3 & 0.9 \\
        Camera, $0.075g$, 2 adjacent, soft & 0.136 & 29.99 & 87.59 & 7.2 & 0.5 & 4.0 & 0.7 \\
        Lattice, no siblings & 0.179 & 39.78 & 84.89 & 9.9 & 0.5 & 4.0 & 0.7 \\
        Lattice, 2 of other tiles & 0.160 & 36.52 & 86.50 & 8.5 & 0.6 & 3.8 & 0.6 \\
        Lattice, 2 random, hard & 0.124 & 24.42 & 87.09 & 7.3 & 0.6 & 4.4 & 0.7 \\
        Lattice, 2 random, soft & 0.142 & 30.45 & 87.33 & 7.6 & 0.5 & 4.0 & 0.6 \\
        LaSo: 2 adjacent, soft & 0.145 & 31.93 & 87.44 & 7.5 & 0.5 & 3.9 & 0.6 \\
        \midrule
        \multicolumn{8}{@{}l}{\emph{DReSS cross}} \\
        Whole tile & 0.454 & 38.58 & 39.30 & 22.5 & 2.1 & 31.2 & 4.9 \\
        Whole tile, 2 of other tiles & 0.442 & 39.22 & 40.06 & 22.3 & 2.2 & 30.8 & 4.7 \\
        Camera, $J = 0$ & 0.071 & 8.02 & 61.94 & 4.3 & 1.4 & 24.2 & 8.3 \\
        Camera, $J = 0.075g$ & 0.117 & 15.40 & 75.88 & 5.5 & 0.8 & 13.2 & 4.6 \\
        Camera, $J = 0.125g$ & 0.159 & 22.27 & 73.70 & 7.7 & 0.8 & 13.4 & 4.4 \\
        Camera, $J = 0.375g$$^\ast$ & 0.370 & 40.41 & 49.20 & 19.6 & 1.7 & 25.0 & 4.5 \\
        Camera, $J = 0.625g$$^\ast$ & 0.548 & 30.60 & 29.96 & 22.7 & 2.8 & 39.4 & 5.1 \\
        Camera, $0.075g$, 2 adjacent, soft & 0.121 & 17.98 & 77.70 & 5.1 & 0.8 & 12.3 & 4.1 \\
        Lattice, no siblings & 0.136 & 18.48 & 75.38 & 6.5 & 0.8 & 13.0 & 4.4 \\
        Lattice, 2 of other tiles & 0.127 & 17.87 & 76.61 & 5.8 & 0.8 & 12.6 & 4.1 \\
        Lattice, 2 random, hard & 0.103 & 13.22 & 76.88 & 4.9 & 0.8 & 13.1 & 4.3 \\
        Lattice, 2 random, soft & 0.124 & 17.66 & 77.46 & 5.4 & 0.8 & 12.3 & 4.0 \\
        LaSo: 2 adjacent, soft & 0.130 & 18.74 & 77.25 & 5.4 & 0.8 & 12.6 & 4.0 \\
        \bottomrule
    \end{tabular}}
    \caption{\textbf{Plateau radius, R@1 and misses over the training positive.} Plateau radius $\rho_{05}$ as in \cref{sec:supp_simabl}, R@1 on the standard gallery ($n = 1$) and on the $5\times5$ index, and the misses on that index in \% of all queries, by type as in \cref{tab:supp_errtypes}. ``2 of other tiles'': two crops of other tiles per training pair as additional negatives, so that the batch holds as many references as with two siblings. Five-epoch fine-tunes of the ViT-B with CC pre-training as in \cref{tab:ablation}, mean over three seeds with a standard deviation of at most 0.63 points, $^\ast$one seed.}
    \label{tab:supp_jmiss}
\end{table}

\section{Number of references and of siblings}
\label{sec:supp_siblings}
\paragraph{References per batch.} With $M = 2$ siblings, a batch holds three times as many references as without. The rows ``2 of other tiles'' of \cref{tab:supp_jmiss} separate the two effects. They add to every training pair two crops of other tiles instead of siblings. They are drawn from the same neighbor lists as the batch, each tile appears at most once per batch, and they are negatives for every query. On the lattice positive, these references recover 1.6 of the 2.5 points by which LaSo exceeds the arm without siblings on VIGOR, and 1.2 of 1.9 on DReSS. The siblings add the remaining 0.6--0.9 points and the better position: the median error of the soft-argmax is 1.81\,m and 4.73\,m for LaSo against 2.00\,m and 5.22\,m with the extra references. On the whole tile, the same references add 0.3 and 0.8 points, and its misses stay those of the whole tile. A larger batch thus does not replace the positive.

\paragraph{Siblings of a camera-centered positive.} The row ``Camera, $0.075g$, 2 adjacent, soft'' of \cref{tab:ablation,tab:supp_ablation,tab:supp_jmiss} adds siblings to the camera-centered positive with $J = 0.075g$. They are two of the eight crops whose centers lie one stride from the center of this positive, drawn at random, cut from the mosaic and not restricted to the tile, with the weights of \cref{eq:sib_weights}. With hard instead of soft targets, this positive reaches 86.41 and 72.96 on the $5\times5$ index, 1.2 and 4.7 points less.

\paragraph{Number and weight of the siblings.} We vary the number $M$ of siblings and the width $\gamma$ of their weights (\cref{eq:sib_weights}) in five-epoch fine-tunes as in \cref{tab:ablation}, with one seed. With $M = 1$, 2, 4 and 8, R@1 on the $5\times5$ index is 87.36, 87.49, 87.65 and 87.56 on VIGOR cross and 76.91, 77.15, 77.76 and 77.88 on DReSS cross, at most 0.7 points from $M = 2$. A width of a quarter of a stride lowers it by 0.5 and 1.2 points, to 86.98 and 75.99. A width of one stride lowers it by 4.9 and 9.6 points, to 82.62 and 67.57, and raises the neighbor misses from 7.5\% and 5.4\% to 12.0\% and 10.3\%. The runs with $M = 4$ and 8 use 16 GPUs with 8 pairs each instead of 8 GPUs with 16 pairs, at the same batch size.

\section{Lattice of training and index}
\label{sec:supp_lattice}
\Cref{tab:supp_lattice} trains LaSo with $n = 3$, 5 and 7 and evaluates every model on all three indexes. For every index, the model trained on the same lattice reaches the highest R@1, with one tie. With matched training and index, R@1 rises by 2.5 points on VIGOR and 7.7 on DReSS from $n = 3$ to 5 and by 0.5 and 2.8 from 5 to 7, and the median error of the soft-argmax falls from 2.48 to 1.81 to 1.58\,m and from 8.65 to 4.73 to 3.55\,m. Each step enlarges the index, to 9, 25 and 49 crops per tile, \ie, 0.42, 1.16 and 2.28\,M crops on VIGOR cross and 1.68, 4.65 and 9.12\,M on DReSS cross. An index coarser than the training lattice costs R@1, up to 7.9 points on VIGOR and 22.8 on DReSS for the model trained with $n = 7$ on the $3\times3$ index. A finer index does not: the model trained with $n = 5$ gains 0.1 and 1.5 points on the $7\times7$ index. The plateau radius $\rho_{05}$ of the three models, one seed each, is 0.217, 0.145 and 0.114\,$g$ on VIGOR and 0.197, 0.129 and 0.097\,$g$ on DReSS.

\paragraph{Finer index without retraining.} The same holds for one seed of the ViT-B with public initialization of \cref{tab:main}, trained with $n = 5$: on indexes with $n = 5$, 7 and 9, its R@1 on VIGOR cross is 85.96, 86.18 and 86.02, and on DReSS cross it rises from 76.10 to 77.25 at $n = 7$. The whole-tile model loses 0.4 and 0.9 points over the same steps.
\begin{table}[h]
    \centering
    \small
    \setlength{\tabcolsep}{3pt}
    \resizebox{\columnwidth}{!}{%
    \begin{tabular}{@{}l ccc ccc@{}}
        \toprule
        & \multicolumn{3}{c}{R@1 on index} & \multicolumn{3}{c}{Soft-argmax [m] on index} \\
        \cmidrule(lr){2-4}\cmidrule(lr){5-7}
        Trained with & $n{=}3$ & $n{=}5$ & $n{=}7$ & $n{=}3$ & $n{=}5$ & $n{=}7$ \\
        \midrule
        \multicolumn{7}{@{}l}{\emph{VIGOR cross}} \\
        $n = 3$ & \textbf{85.00} & 84.66 & 83.99 & 2.48 & 2.50 & 2.76 \\
        $n = 5$ & \textbf{85.00} & \textbf{87.49} & 87.57 & 2.75 & 1.81 & 1.78 \\
        $n = 7$ & 80.08 & 87.21 & \textbf{87.99} & 3.48 & 1.75 & 1.58 \\
        \midrule
        \multicolumn{7}{@{}l}{\emph{DReSS cross}} \\
        $n = 3$ & \textbf{69.47} & 71.33 & 71.69 & 8.65 & 7.41 & 7.71 \\
        $n = 5$ & 68.50 & \textbf{77.15} & 78.64 & 10.18 & 4.73 & 4.28 \\
        $n = 7$ & 57.13 & 76.54 & \textbf{79.97} & 14.68 & 5.08 & 3.55 \\
        \bottomrule
    \end{tabular}}
    \caption{\textbf{Training lattice against index lattice.} R@1 and median localization error of the soft-argmax for LaSo trained with $n = 3$, 5 and 7 (rows) on indexes with $n = 3$, 5 and 7 (columns). Five-epoch fine-tunes of the ViT-B with CC pre-training as in \cref{tab:ablation}, one seed. Best R@1 per index in bold.}
    \label{tab:supp_lattice}
\end{table}

\end{document}